\documentclass[num-refs]{nbdt-article}
\papertype{Original Article}
\paperfield{Analysis}

\usepackage[utf8]{inputenc}
\usepackage{natbib}
\usepackage{amssymb}
\usepackage{amsmath}
\usepackage{graphicx}
\usepackage{color}
\usepackage{bm}
\usepackage{nicefrac}
\usepackage{appendix}
\usepackage{float}
\usepackage{anyfontsize}
\pdfoutput=1

\graphicspath{{figures/}}

\definecolor{movecol}{rgb}{0,0.5,0}
\newcommand{\edit}[1]{{#1}}

\newcommand{\balpha}{\boldsymbol{\alpha}}

\title{Correlated Components Analysis -- Extracting Reliable Dimensions in Multivariate Data}

\author[1]{Lucas C. Parra}
\author[2]{Stefan Haufe}
\author[1]{Jacek~P.~Dmochowski}

\affil[1]{Department of Biomedical Engineering,
      City University of New York,
      New York, NY 10031, USA}
\affil[2]{Berlin Center for Advanced Neuroimaging,
      Charit\'e - Universit\"atsmedizin Berlin,
      10117 Berlin, Germany}

\corraddress{Department of Biomedical Engineering,
      City University of New York,
      New York, NY 10031, USA}
\corremail{parra@ccny.cuny.edu}

\fundinginfo{This work was supported through grants from NSF (DRL-1660548) and ARL (W911NF-10-2-0022).}

\runningauthor{Parra, Haufe, Dmochowski}

\begin{document}

\maketitle

\begin{abstract}%
How does one find dimensions in multivariate data that are reliably expressed across repetitions? For example, in a brain imaging study one may want to identify combinations of neural signals that are reliably expressed across multiple trials or subjects.  For a behavioral assessment with multiple ratings, one may want to identify an aggregate score that is reliably reproduced across raters.  Correlated Components Analysis (CorrCA) addresses this problem by identifying components that are maximally correlated between repetitions (e.g. trials, subjects, raters). Here we formalize this as the maximization of the ratio of between-repetition to within-repetition covariance. We show that this criterion maximizes repeat-reliability, defined as mean over variance across repeats, and that it leads to CorrCA or to multi-set Canonical Correlation Analysis, depending on the constraints. Surprisingly, we also find that CorrCA is equivalent to Linear Discriminant Analysis for \edit{zero}-mean signals, which provides an unexpected link between classic concepts of multivariate analysis. We present an exact parametric test of statistical significance based on the F-statistic for normally distributed independent samples, and present and validate shuffle statistics for the case of dependent samples. Regularization and extension to non-linear mappings using kernels are also presented. The algorithms are demonstrated on a series of data analysis applications, and we provide all code and data required to reproduce the results.

% Please include a maximum of seven keywords
\end{abstract}

\section{Introduction}

Consider the following scenario: A group of people are shown a movie while their neural activity is recorded.  Guided by the assumption that the movie evokes similar brain responses in different subjects, the goal is to identify movie-related brain activity that is common across individuals  \citep{dmochowski2012correlated}. Because activity is distributed across multiple sensors, we would like to find linear combinations of sensors that have a similar time course in all subjects. In other words, we would like to identify underlying \emph{components} or factors that have high \emph{inter-subject correlation}.

In a different setting, the task may be to assess motor skills. A clinician observes a group of individuals \edit{performing} various tasks and provides a rating for each person and task  \citep{rousson2002assessing}. Typically, ratings are combined across tasks so that each person obtains a single aggregate performance score. But how should individual task ratings be combined? Traditionally, this is done by simple averaging of related scores. We propose instead a data-driven approach that aims to make the aggregate scores as consistent as possible across different raters. This means that we want to combine scores so as to maximize \emph{inter-rater reliability}. 

What is common in these two scenarios is the goal of identifying directions in multivariate data sets that are reliably reproduced across repetitions.  Correlated Components Analysis (CorrCA) accomplishes this goal. CorrCA was originally developed in the context of neuroimaging studies to extract similar activity in multiple subjects \citep{dmochowski2012correlated}, and was re-developed independently to extract reliable brain responses across repeated trials in a single subject \citep{tanaka2013task}. Here we will show that the technique can also be used to identify aggregate ratings with high inter-rater reliability. 

More generally, CorrCA is applicable whenever a data volume is available with dimensions $T \times D \times N$, where $N$ are \edit{repeated measures with the same sensors}. The method identifies directions in the $D$-dimensional space along which the \edit{data} maximally correlate between $N$ repeated measurements, with correlation measured across $T$ samples. This is similar to Principal Components Analysis \citep[PCA,][]{hotelling1933analysis} in that CorrCA finds a set of $D$-dimensional projection vectors that linearly decompose a data set (of size $T \times D$).  Instead of capturing dimensions with maximum variance within a single data set as in PCA, CorrCA captures the dimensions with maximum correlation between the $N$ data sets. As we will show here, the optimality criterion of CorrCA can also be used to derive Canonical Correlation Analysis \citep[CCA,][]{hotelling1936relations}, including multi-set CCA \citep[MCCA,][]{mckeon1966canonical,kettenring1971canonical}. The methods differs in that CorrCA uses a single set of linear projections for all data sets \edit{assuming they repeated measures are all taken in the same space}, whereas CCA and MCCA yield a different set of projection vectors for each \edit{data-set}. MCCA is well established for the purpose of identifying similar brain activity across subjects \citep{li2009joint,correa2010canonical,hwang2012functional,afshin2012enhancing,lankinen2014intersubject,zhang2017inter}. Here we will show for brain signals and rating data that CorrCA can achieve better performance than MCCA by virtue of the smaller number of free parameters.

The main contribution of this paper is to provide a \edit{didactic yet} formal characterization of CorrCA and with this, establish a few novel theoretical results. In particular, we show that maximizing the correlation between repeated measurements is indeed equivalent to maximizing the reliability of the projected data. By ``reliability'' we mean the average over repetitions, divided by the variance over repetitions, which is also a sensible definition for signal-to-noise ratio. We establish a direct link to MCCA, as well as Linear Discriminant Analysis (LDA). CorrCA and MCCA both maximize inter-subject correlation, defined here the ratio of between- to within-subjects covariance, while LDA  maximizes the between-class over the within-class variance. Surprisingly, LDA and CorrCA \edit{yield} the same result for \edit{zero}-mean signals. This direct equivalence allows us to formulate a strict statistical test for significance based on Fisher's variance ratio (F-statistic). The equivalence to LDA suggests a straightforward extension of CorrCA to non-linear mappings using the well-known ``kernel'' approach. We also provide a validation of CorrCA for identifying shared dimensions using simulated data and validate the proposed tests of statistical significance. Another important contribution of this paper is to formalize inter-subject correlation (ISC) as a correlation metric which applies to more than two signals. ISC differs from conventional definitions of correlation, yet seamlessly integrates into the frameworks of Linear Discriminant Analysis, Canonical Correlation Analysis, and the F-statistic.

Before we present the mathematical treatment, we begin with a simple two-dimensional example to \edit{illustrate} the basic concept of identifying reliable signal components that are contained in a shared linear dimension. 

\begin{figure}[htbp]
\begin{center}
\includegraphics[width=0.9\columnwidth]{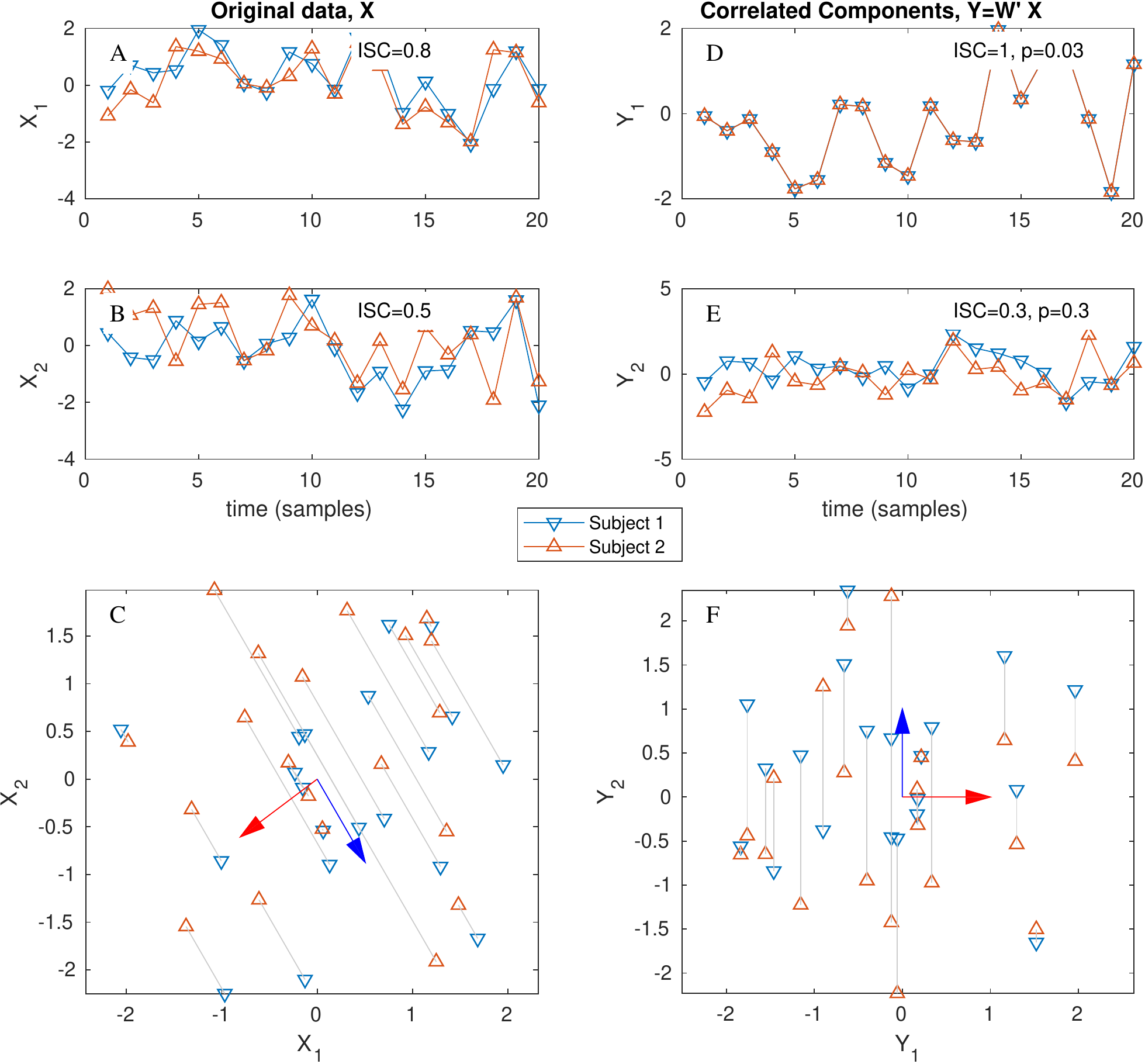}
\end{center}
\caption{{\bf Extraction of a correlated component \edit{from} 2D time courses.} A common electrical current source, $s(t)$, is captured by voltage sensors $x_{1l}(t)$ and $x_{2l}(t)$ for subject $l$. This common source contributes to both subjects with identical scaling $a$ and $b$: $x_{1l}(t)=a s(t)+ c \xi_l(t)$, $x_{2l}(t)=b s(t)+ d\xi_l(t)$, where $\xi_l(t)$ is a source that contributes a different value to each subject $l$ but with the same mixing coefficients $c, d$.  Note that \edit{voltages from multiple} current sources \edit{combine} additively in a resistive medium. (A) Time courses $x_{1l}(t)$ for both subjects. (B) Time courses $x_{2l}(t)$ (C) Same data, but now individual time points are connected (by gray line) between the subjects.  The source directions $[a, b]$ and $[c, d]$ are indicated with a red and blue arrow respectively and are the same for both subjects, i.e. they are ``shared'' dimensions.  (D) Projection $y_{1l}(t)$ with maximal correlation. The time courses of the two subjects are perfectly correlated with inter-subject correlation ISC=1. (E) Projection $y_{2l}(t)$ that is not common between subjects. (F) Along dimension $y_1$, both subjects' data are expressed identically, i.e. this is the  (``reliable'') source dimension that is shared between the two subjects. Values along dimension $y_2$ differ randomly between subjects. This dimension is not ``reliable'' across subjects. }
\label{fig:toy-example-2D} 
\end{figure}

\subsection*{Extracting shared dimensions --- an illustrative example}

For this illustrative example we consider a simulation of electrical brain signals recorded from multiple subjects. Assume that we have two sensors, $x_1(t)$ and $x_2(t)$, which record \edit{temporal} signals from each of two subjects, as shown in Figure~\ref{fig:toy-example-2D}~A \& B. The figure caption details the generation of the data; briefly, there are two ``sources'' of signal: one has a common waveform for both subjects, while a second source has a subject-dependent waveform. The common waveform \edit{indroduces} a correlation between subjects at both sensors, although the exact shared waveform is not immediately apparent in the individual time courses shown in panels A \& B). The source directions become clear from a scatter plot of the same data (panel~C, blue and red arrows). Here, for each time point $t$ we have connected the samples of both subjects. Evidently, at any point in time, the signals of the two subjects lie along a line (in the direction of the source not shared by the subjects, blue arrow), but the lines are shifted in parallel (in direction of the common source, red arrow). 
The locations along the gray lines are determined by the subject-dependent waveform, while the shift of the parallel line from one time instance to the next is determined by the common waveform. 
\edit{It is important to note that while both directions are shared by the two subjects, only one of these dimensions expresses a common waveform.} 
Now we select a linear transformation, ${\bf V}$, to map this data onto a component space, ${\bf y}(t)={\bf V}^\top {\bf x}(t)$,
such that the component signals are maximally correlated between subjects. With this choice, the time course for both subjects in the first component dimension, $y_1(t)$, are now perfectly correlated (panel~D). Thus, we have extracted the common source of variation in these two subjects, whereas the second component dimension, $y_2(t)$, now captures a waveform that is \edit{not necessarily shared} between subjects (panel~E). As a result, the gray lines in the scatter plot are now vertical (panel~F), separating the \edit{waveforms} into reproducible \edit{signal} ($y_1$) and non-reproducible \edit{noise} dimensions ($y_2$). Evidently, ${\bf V}$ captured the source directions in these data (blue and red arrows). The objective of Correlated Components Analysis is to identify the unknown projection matrix ${\bf V}$ by analyzing time courses ${\bf x}(t)$. \edit{The fundamental assumption of CorrCA is that there is a reproducible signal and a non-reproducible noise, and that the directions of the reproducible signal are shared between subjects.} 

\edit{Readers wishing to omit the theoretical treatment below may proceed directly to Section \ref{sec:apps}, which presents several applications of CorrCA to real-world data sets.}  

\section{Theory}

\subsection{Problem statement: Maximal inter-subject correlation}
\label{sec:max-isc}

The objective of CorrCA is to find dimensions in multivariate data that maximize correlation between repetitions. Denote the $D$-dimensional data by ${\bf x}_i^l \in \mathbb{R}^D$, where $i=1, \dots ,T$ enumerates exemplars, and $l=1, \dots, N$ represents repeated renditions of these data. In the case of brain signals, the exemplars represent time points, so that $i$ indexes a sequence in time of length $T$, and $l$ indexes $N$ repeated recordings (subjects or trials). In the case of a behavioral assessment instrument, $i$ may index individuals receiving a rating, while $l$ indexes multiple raters, or repeated ratings given to the same individuals by a single rater. In both instances, the observations are $D$-dimensional ($D$ sensors that record brain signals, or $D$ items rated in the assessment instrument). For simplicity, in the following we will \edit{employ the terms ``time'' and ``subjects'' in reference to $i$, such} that correlation is measured \emph{between subjects} and \emph{across time}. When appropriate, we will rephrase the resulting expressions for the case of assessing inter-rater correlation or test-retest correlation. 

The goal is now to identify a linear combination of these $D$ measures, defined by a projection vector ${\bf v} \in {\mathbb{R}}^D$:
\begin{equation}
y^l_i = {\bf v}^\top {\bf x}^l_i , 
\label{projection}
\end{equation} 
such that the correlation between $N$ subjects (repetitions) is maximized.  We define inter-subject correlation (ISC) as the ratio of the between-subject covariance, $r_B$, and the within-subject covariance, $r_W$: 
\begin{equation}
\rho = \frac{1}{N-1}\frac{r_B}{r_W} .
\label{isc}
\end{equation}
The between-subject covariance is a sum over all pairs of subjects, and within-subject covariance is a sum over all subjects: 
\begin{eqnarray}
r_B &=& \sum_{l=1}^N \sum_{k=1, k \neq l}^N \sum_{i=1}^T   (y_i^l - {\bar y}_*^l)(y_i^k - {\bar y}_*^k) 
\label{between-subject-covariance} , \\
r_W &=& \sum_{l=1}^N \sum_{i=1}^T (y_i^l - {\bar y}_*^l) (y_i^l - {\bar y}_*^l)
\label{within-subject-covariance} ,
\end{eqnarray}
where, $\bar{y}^l_* =  \frac{1}{T} \sum_{i=1}^T y_i^l$, is the sample-mean \edit{of the projected data} for subject $l$. The factor $(N-1)^{-1}$ in definition (\ref{isc}) is required so that correlation is normalized to $\rho \le 1$ (see Appendix~\ref{apdx:normalization-of-isc}). To simplify the notation, in our ``variance'' and ``covariance'' definitions we generally omit common scaling factors, in this case \edit{we omitted} $(T-1)^{-1}N^{-1}$.

While we refer to $\rho$ as the inter-subject correlation (ISC), it could also be termed the inter-rater correlation or inter-repeat correlation (IRC) depending on the context. In Appendix~\ref{apdx:ICC} we discuss this definition of correlation and its relation to other measures of correlation, such as intra-class correlation or Pearson's correlation coefficient. Note that $r_W$ is also the variance around the subject-mean.  

By inserting Eq. (\ref{projection}) into definition (\ref{isc}) it follows readily that
\begin{equation}
\rho = \frac{1}{N-1} \frac {{\bf v}^\top {\bf R}_B {\bf v} }{ {\bf v}^\top {\bf R}_W {\bf v}} ,
\label{isc-x}
\end{equation}
where ${\bf R}_B$ and ${\bf R}_W$ are the between-subject and within-subject covariance matrices of ${\bf x}_i^k$ defined analogously to (\ref{between-subject-covariance})-(\ref{within-subject-covariance}):
\begin{eqnarray}
{\bf R}_B &=& 
\sum_{i=1}^T \sum_{l=1}^N \sum_{k=1, k \neq l}^N 
({\bf x}_i^l - {\bar {\bf x}}_*^l)({\bf x}_i^k - {\bar {\bf x}}_*^k) ^\top ,
\label{between-subject-covariance-matrix} \\
{\bf R}_W &=& 
\sum_{i=1}^T \sum_{l=1}^N 
({\bf x}_i^l - {\bar {\bf x}}_*^l)({\bf x}_i^l - {\bar {\bf x}}_*^l)^\top .
\label{within-subject-covariance-matrix}
\end{eqnarray}
Here ${\bar {\bf x}}_*^l = \frac{1}{T} \sum_{i=1}^T {\bf x}_i^l$ is the sample-mean vector for subject $l$. The projection vector ${\bf v}$ that maximizes $\rho$ can be found as the solution of  $\partial \rho / \partial {\bf v}^\top = 0$, which yields: 
\begin{equation}
{\bf R}_B {\bf v} = \rho \, {\bf R}_W {\bf v}   .
\label{eigenvalue-equation-isc}
\end{equation}

\edit{Assuming} that ${\bf R}_W$ is invertible, ${\bf v}$ is an eigenvector of ${\bf R}_W^{-1} {\bf R}_B$ with eigenvalue $\rho$. Maximal ISC is achieved by projecting the data onto the eigenvector of ${\bf R}_W^{-1} {\bf R}_B$ with \edit{largest} eigenvalue. We refer to this principal eigenvector as ${\bf v}_1$. In addition to this projection with maximal ISC, there may be additional dimensions that capture significant ISC in the data. Our goal is to find all such projection vectors, ${\bf v}_d, d=1 \ldots D$. Thus, \edit{a natural objective} is to maximize ISC summed over all components. However, we want each component to capture a different, independent aspect of the data. We therefore require components to be uncorrelated, similar to the constraints enforced by Independent Components Analysis or Canonical Correlation Analysis. (The alternative of requiring orthogonal projection vectors, as in PCA is discussed in Appendix \ref{apdx:uncorrelated-vs-orthogonal}.)  Thus, in total the goal is to maximize the following cost function with respect to ${\bf V} = [{\bf v}_1, \ldots {\bf v}_K]$:
\begin{equation}
J({\bf V}) = \sum_{d=1}^{D} \rho_d = \sum_{d=1}^{D} 
\frac
{{\bf v}_d^\top {\bf R}_B {\bf v}_d}
{{\bf v}_d^\top {\bf R}_W {\bf v}_d}
\label{summed-isc}
\end{equation}
subject to the constraints that components are mutually uncorrelated, i.e., ${\bf v}_d^\top {\bf R}_W {\bf v}_c =0$, for all $c \ne d$. The solution to this problem is given by the  eigenvector matrix \edit{${\bf V}$} that satisfies the eigenvalue equation (see Appendix~\ref{apdx:max-sum-isc}):
\begin{equation}
{\bf R}_B {\bf V} =  {\bf R}_W {\bf V} {\boldsymbol \Lambda},
\label{eigenvalue-equation-matrix}
\end{equation}
where ${\boldsymbol \Lambda}$ is a diagonal matrix with diagonal terms measuring the ISC, $\rho_d$, of the corresponding projections of the data, $y_d = {\bf v}_d^\top {\bf x}$. Note that these additional projections of the data are uncorrelated, as desired, because their correlation matrix, ${\bf R}^{y}_W$  --- defined as in Eq. (\ref{within-subject-covariance-matrix}) but for ${\bf y}$ --- is diagonal:

\begin{equation}
{\bf R}^{y}_W = {\bf V}^\top {\bf R}_W {\bf V} =  {\boldsymbol \Lambda}_W .
\label{diagonalization-Rw}
\end{equation}
The fact that ${\boldsymbol \Lambda}_W$ is a diagonal matrix is a well-know property of the generalized eigenvalue equation (\ref{eigenvalue-equation-matrix}) (see Appendix~\ref{apdx:joint-diagonalization}).

To summarize, the principal eigenvector of the between-subject over the within-subject covariance matrices provides the projection of the data that maximizes ISC. Subsequent smaller eigenvectors provide additional projections of the data that are uncorrelated with one another. The corresponding eigenvalues are equal to the ISC that each projection achieves. It follows that the last projection captures the direction with the smallest ISC. Note that the corresponding \edit{eigenvectors} are uniquely defined, provided that the ISC are distinct (see Appendix~\ref{apdx:identifiability}). 

In the illustrative example of Figure~\ref{fig:toy-example-2D} with $D=2$, the eigenvalues were $\rho_1=1$ and $\rho_2=0.04$, for the red and blue directions respectively (see Figure~\ref{fig:toy-example-2D}~D and \ref{fig:toy-example-2D}~F). Evidently the shared dimension where the two subjects vary in unison (red) has been identified. A concrete application of CorrCA to the problem of identifying brain activity that is shared between subjects is described in section~\ref{app:EEG-video-example}. A thorough evaluation of the conditions under which the algorithm \edit{identifies the correct} projection vectors ${\bf v}_d$ will be presented in  section~\ref{sec:identification-snr} (Fig.~\ref{fig:sim-SNR-dependence}). \edit{The main finding is that the shared waveforms (signal) need to project in the same direction across subjects. Performance degrades if there is additive noise, or equivalently, if the directions of the non-reproducible waveforms (noise) differ between subjects.}  

Given a set of extracted correlated components, the question arises as to how many of these components exhibit statistically significant correlations. For zero mean, independent test data we can use the F-statistic to determine statistical significance (see section~\ref{sec:maximum-reliability}, Eq.~\ref{F-statistic}). Statistical testing is described in Appendix~\ref{sec:significance}, where we also treat the case of non-independent data using shuffle statistics. The methods for statistical testing will be validated with numerical simulations in section~\ref{sec:simulations}.  

\subsection{Fast computation of inter-subject correlation}
\label{sec:speedup}
In Eq. (\ref{between-subject-covariance-matrix}), the term $k=l$ is excluded from the sum over $k$. Note that the excluded term is precisely ${\bf R}_W$, so that adding the two covariances together completes a sum with all pairs, including $k=l$. We will refer to this, therefore, as the total covariance:
\begin{eqnarray}
{\bf R}_T &=& {\bf R}_B + {\bf R}_W 
\label{covariance-additivity} \\
&=&
\sum_{i=1}^T \sum_{l=1}^N \sum_{k=1}^N 
({\bf x}_i^l - {\bar {\bf x}}_*^l)({\bf x}_i^k - {\bar {\bf x}}_*^k) ^\top .
\end{eqnarray}
This relationship is useful because ${\bf R}_T$ can be simplified to: 
\begin{equation}
{\bf R}_T = N^2 \sum_{i=1}^T  
({\bar {\bf x}}_i^* - {\bar {\bf x}}_*^*)({\bar {\bf x}}_i^* - {\bar {\bf x}}_*^*) ^\top ,
\label{total-covariance-matrix}
\end{equation}
where ${\bar {\bf x}}_i^* = \frac{1}{N} \sum_{l=1}^N {\bf x}_i^l$ and ${\bar {\bf x}}_*^* = \frac{1}{T} \sum_{i=1}^T {\bar {\bf x}}_i^*$ are the mean across subjects and the grand mean, respectively.  Thus, surprisingly, to compute inter-subject correlation, one never has to \edit{explicitly correlate} pairs of subjects, because ${\bf R}_B$ can be computed from ${\bf R}_T$ and ${\bf R}_W$, and neither one requires a sum over pairs of subjects. With an order of $N$ operations, this makes the computation more efficient than the direct implementation of Eqs.~(\ref{between-subject-covariance-matrix})-(\ref{within-subject-covariance-matrix}), which requires an order of $N^2$ operations. The direct implementation was used in all previous work \citep[e.g.,][]{dmochowski2012correlated,tanaka2013task,cohen2016memorable}, but becomes problematic when one uses populations sizes reaching hundreds of subjects or more \citep[e.g.][]{petroni2018variability,Alexander149369}. 

\subsection{Relationship to Linear Discriminant Analysis}
\label{sec:LDA}
 
Perhaps the naming and the additivity of the covariance matrices introduced here --- Eqs. (\ref{between-subject-covariance-matrix}), (\ref{within-subject-covariance-matrix}), (\ref{covariance-additivity}) --- reminds some readers of the between-class and within-class scatter matrices used in Linear Discriminant Analysis (LDA). In the current notation these are:
\begin{eqnarray}
{\bf S}_T &=& 
\sum_{i=1}^T \sum_{l=1}^N  
({\bf x}_i^l - {\bar {\bf x}}_*^*)({\bf x}_i^l - {\bar {\bf x}}_*^*) ^\top 
\label{total-scatter-matrix}\\
{\bf S}_B &=&
\sum_{i=1}^T   N
({\bar {\bf x}}_i^* - {\bar {\bf x}}_*^*)({\bar {\bf x}}_i^* - {\bar {\bf x}}_*^*) ^\top 
\label{between-scatter-matrix}\\
{\bf S}_W &=&
\sum_{i=1}^T \sum_{l=1}^N  
({\bf x}_i^l - {\bar {\bf x}}_i^*)({\bf x}_i^l - {\bar {\bf x}}_i^*) ^\top .
\label{within-scatter-matrix}
\end{eqnarray}
Here, index $i$ enumerates the different classes and index $l$ enumerates the exemplars in each class (see Table~\ref{table-indice} in \edit{Appendix~\ref{apdx:ICC}} for an overview on the use of indices in different algorithms). Therefore $\bar{{\bf x}}_i^* =  \frac{1}{N} \sum_{l=1}^N \bar{{\bf x}}_i^l$ and $\bar{{\bf x}}_*^* = \frac{1}{T} \sum_{i=1}^T \bar{\bf {x}}_i^*$ are the class mean and grand mean vectors, respectively. Scatter matrices also satisfy an additivity rule: ${\bf S}_T={\bf S}_B +{\bf S}_W $ \citep{duda2012pattern}. The goal of LDA is to maximize the separation between classes, which is \edit{defined as} the ratio of between- to within-class variance: 
\begin{equation}
S = \frac {{\bf v}^\top {\bf S}_B {\bf v} }{ {\bf v}^\top {\bf S}_W {\bf v}} .
\label{Separation}
\end{equation}
As before, the maximal separation $S$ is obtained with the eigenvector of ${\bf S}_B^{-1}{\bf S}_W$ \edit{corresponding to} the largest eigenvalue $S$. As before, we can extract additional projections that \edit{achieve separation of classes}. If we require that these projections are uncorrelated, then they are given by the additional eigenvectors of ${\bf S}_B^{-1}{\bf S}_W$. This concept was first proposed by \cite{rao1948utilization} and generalizes the case of two classes introduced by \cite{fisher1936use}. We refer to this as classical LDA, and note that it maximizes the sum of variance ratios $S$ computed separately for each projection (see ~\ref{apdx:max-sum-isc}). An alternative approach is needed if instead the goal is to maximize the ratio of summed variances \citep{yan2006trace}, or if one prefers orthogonal projection \citep{cunningham2015linear}. 

This optimality criterion and resulting eigenvalue problem look strikingly similar to that of CorrCA in (\ref{isc-x}). Despite the similar naming, however, the scatter matrices differ from the between-subject and within-subject covariance matrices, and it is not immediately obvious how the two optimality criteria relate to one another. There is, however, a close relationship between the two (see Appendix~\ref{apdx:scatter-vs-covariance}): 
\begin{eqnarray}
N {\bf S}_B &=& {\bf R}_W +  {\bf R}_B , \\
N {\bf S}_W &=& (N-1) {\bf R}_W - {\bf R }_B   + N T {\bf S}_M ,   
\end{eqnarray}
where ${\bf S}_M$ is the covariance of $\bar{\bf x}^l_*$ across subjects:
\begin{equation}
{\bf S}_M = 
 \sum_{l=1}^N ({\bar {\bf x}}_*^l - {\bar {\bf x}}_*^*)
              ({\bar {\bf x}}_*^l - {\bar {\bf x}}_*^*)^\top .
\label{variance-of-mean}     
\end{equation}
When these \edit{sample}-mean \edit{vectors} are equal across subjects then the scatter matrices are a linear combination of the between- and within-subject covariance matrices. \edit{Note that equal sample-means, ${\bar {\bf x}}_*^l={\bar {\bf x}}_*^k$, does not imply equal class-means, ${\bar {\bf x}}_i^*={\bar {\bf x}}_j^*$. Zero-mean signals will naturally satisfy this condition and will typically still have different class means.} Under this assumption, we can also write a simple functional relationship between class separation $S$ and inter-subject correlation $\rho$:
\begin{equation}
S = \frac{r_W+r_B}{(N-1) r_W - r_B} = \frac{\rho+(N-1)^{-1}}{1-\rho}. 
\label{ISCtoSNR}
\end{equation}
\edit{This relationship will be useful in section~\ref{sec:maximum-reliability} where we establish a link between ISC and the classic F-statistic (Eq.~\ref{F-statistic}), which yields a parametric test for statistical significance of ISC.  Note that $S$ is monotonically increasing with $\rho$} (for $\rho\le 1$), because the slope of this relationship is strictly positive: 
\begin{equation}
\frac{\partial S}{\partial \rho} = \frac{N}{N-1}\frac{1}{(1-\rho)^2} > 0.
\end{equation}
This means that maximizing class separation $S$ in (\ref{Separation}) also results in maximal ISC $\rho$, provided that we equate the means. Thus, finding vectors ${\bf v}$ that maximize $S$ of Eq. (\ref{Separation}) gives the same set of solutions as maximizing $\rho$ in Eq. (\ref{isc-x}). 

To see this, note that the goal of maximizing a ratio of two matrices, such as in Eq. (\ref{isc-x}), can also be achieved by joint-diagonalization of the two matrices (see Appendix~\ref{apdx:joint-diagonalization}):
\begin{equation}
\begin{split}
{\bf V}^\top {\bf R}_B {\bf V}  &= {\boldsymbol \Lambda}_B , \\
{\bf V}^\top {\bf R}_W {\bf V}  &= {\boldsymbol \Lambda}_W ,
\end{split}
\label{joint-diagonalization-Rb-Rw}
\end{equation}
where ${\boldsymbol \Lambda}_B$ and ${\boldsymbol \Lambda}_W$ are diagonal matrices. Note that if ${\bf V}$ jointly diagonalizes two matrices, it also diagonalizes any linear combinations of the two \citep[briefly recapitulated in Appendix~\ref{apdx:linear-combinations}]{fukunaga2013introduction}. Because ${\bf S}_B$ and ${\bf S}_W$ are linear combination of ${\bf R}_B$ and ${\bf R}_W$, then the ${\bf V}$ that satisfies (\ref{joint-diagonalization-Rb-Rw}) also diagonalizes ${\bf S}_B$ and ${\bf S}_W$.  Thus, this ${\bf V}$ captures the eigenvectors to ${\bf R}_W^{-1} {\bf R}_B$ as well as ${\bf S}_W^{-1} {\bf S}_B$, and satisfies in both cases the constraint of mutually uncorrelated components. Because the  eigenvalues of the two problems are monotonically related, the eigenvectors will be sorted in the same order for both problems. Thus, the solutions to both optimization problems are the same, including the constraint of uncorrelated components. To summarize, CorrCA and classical LDA yield the same result, provided that the sample mean vectors are equal across subjects.

The equivalence between CorrCA and LDA is perhaps surprising. \edit{Note that LDA attempts to separate classes so that exemplars of one class do not overlap with the exemplars of other classes. In the illustrative example of Figure~\ref{fig:toy-example-2D}, time samples $i$ take on the role of classes. There are 20 classes, each with two exemplars (connected with gray lines). Classes are well separated in the direction of component $y_1$ (red arrow), whereas they are overlapping in the original dimensions $x_1$ and $x_2$.  Note that the signals are zero-mean across samples but have differing class-means. In this toy example, there is no variation between the two exemplars in each class (vertical lines in panel F indicate that the $y_1$ values are identical within a class). Thus, CorrCA has found a dimension of the data ($y_1$) in which the classes are perfectly separated. Zero within-class variance leads to infinite separation (Eq.~\ref{Separation}) and this in turn to unit correlation  (Eq.~\ref{ISCtoSNR}).  More generally,} we have demonstrated \edit{with Eq.~(\ref{ISCtoSNR})} that increasing separation between samples is equivalent to increasing correlation across samples. 
To understand this, note that separation is high when multiple repeats have similar values, yet these values have to be different for different samples. This means that multiple repeats have to vary {\em together} across samples. In other words, the variations across samples have to be highly correlated across repeats.

\edit{Notice that the equivalence between CorrCA and LDA requires the same number of exemplars in each class -- something that is not usually the case in conventional LDA. In fact, there is a one-to-one link between exemplars across classes (gray lines in panels C and F of Figure~\ref{fig:toy-example-2D}).}

\begin{table}
\caption{Relationship between the optimality criteria of different algorithms. }
\label{table-criteria}
\begin{center}
\begin{threeparttable}
{
\begin{tabular}{lll}
\headrow
\thead{CorrCA} & \thead{LDA} & \thead{JD} \\
$\rho=\frac{1}{N-1}\frac{{\bf v}^\top {\bf R}_B {\bf v}}{{\bf v}^\top {\bf R}_W {\bf v}}$ 
\hspace{1.5cm}
& 
$S=\frac{{\bf v}^\top {\bf S}_B {\bf v}}{{\bf v}^\top {\bf S}_W {\bf v}} \stackrel{*}{=} \frac{\rho+(N-1)^{-1}}{1-\rho}$  
\hspace{1.5cm}
&
$\frac{{\bf v}^\top {\bf S}_B {\bf v}}{{\bf v}^\top {\bf S}_T {\bf v}}\stackrel{*}{=}(N-1)\rho + 1$ 
\hspace{1.5cm}
 \\
\hline
\end{tabular}
}
\begin{tablenotes}
\item $\rho$: inter-subject correlation,  $S$: class-separation. * assuming zero-mean signals, or unbiased raters (${\bf S}_M=0$).
\end{tablenotes}
\end{threeparttable}
\end{center}
\end{table}

\subsection{Maximum reliability and the F-statistic}
\label{sec:maximum-reliability}
In the present context, the scatter matrices ${\bf S}_B$ and ${\bf S}_W$ are interesting for another, perhaps more important reason. Consider the case where $l$ represents repeated measurements of raters, or repeated measures on the same subjects, so that $\rho$ now measures inter-rater correlation, or inter-repeat correlation (IRC). According to Eq. (\ref{between-scatter-matrix}), ${\bf S}_B$ measures  (except for a scaling factor) the sample-covariance of ${\bar {\bf x}}_i^*$, which is the mean across repeats. On the other hand, according to Eq.  (\ref{within-scatter-matrix}), ${\bf S}_W$ measures the sample-covariance averaged over repeats. When projected onto $y$ with ${\bf v}$ they define  the variance of the mean across repeats, $\sigma_{\bar y}^2$, and  average variance around these means, ${\bar \sigma_y}^2$, respectively.  The ratio of the two variances is a sensible definition for signal-to-noise ratio (SNR):
\begin{equation}
S = \frac{\sigma_{\bar y}^2}{\bar \sigma_y^2} .
\label{SNR}
\end{equation} 
We take this SNR also as a metric of repeat-reliability. In this view, the results of the previous section show that maximizing correlation between repeats is equivalent to maximizing repeat-reliability. In particular, Eq. (\ref{ISCtoSNR}) provides the relationship between SNR and ISC. An application of CorrCA to the problem of identifying ratings that are reliably reproduced between different raters is described in Sections~\ref{app:ratings} and \ref{app:HBN-ratings}. In Section~\ref{app:EEG-ERP} we describe results on the problem of identifying components of brain activity that is reliably reproduced across repeated trials. \edit{In all instances, the goal is to maximize $\rho$, which is equivalent to maximizing $S$.}

A component decomposition method that specifically maximizes SNR
was previously described in \citet{de2014joint}. This method can be formulated as a joint diagonalization (JD) problem similar to CorrCA or LDA. When SNR is defined as in Eq.~(\ref{SNR}), then the JD approach diagonalizes matrices ${\bf S}_T$ and ${\bf S}_B$ in the present notation. By the same argument as in section~\ref{sec:LDA}, it is clear that this JD approach provides the same solution as LDA and CorrCA. The optimization criteria for the three techniques are summarized in Table~\ref{table-criteria}.

We note that $S$ is a ratio of variances, which permits a direct link to the $F$-statistic, as used in the analysis of variance:
\begin{equation}
F = \frac{T(N-1)}{T-1} S = \frac{T(N-1) \rho+T}{(T-1)(1-\rho)} .
\label{F-statistic}
\end{equation}  
For normally and independently distributed samples this quantity follows the $F$-distribution with degrees of freedom $d_1=T(N-1)$ and $d_2=T-1$  \citep[Chapter 6]{papoulis2002probability}.  We can therefore use the $F$-distribution to \edit{assess} statistical significance for $\rho$. This gives us a strict parametric test of significance for each CorrCA component (on unseen data and provided the means are equalized, i.e. $\bar{\bf x}^l_*=\bar{\bf x}^*_*$). We will leverage this observation when we perform tests for statistical significance of $\rho$ evaluated on \edit{independently and identically distributed} test data. \edit{For non-independent samples, which are more typical of the temporal signals of interest, we will rely instead on shuffle-statistics, which can also be applied directly to training data (see Appendix~\ref{sec:significance}). Both methods will be evaluated for accuracy in section~\ref{sec:simulations}.} 
 
\subsection{Forward model}
\label{sec:forward-model}

An import aspect of multivariate models is parameter interpretation. CorrCA is an example of a ``backward model'' in the sense that the observed data is transformed into components that capture the source of covariation. It shares this property with a host of other methods such as PCA and ICA. An alternative strategy is ``forward modeling'' such as Factor Analysis or General Linear Models, which transforms known variables to \edit{explain} the observed data \citep{haufe2014interpretation}. Such models often capture the physical processes underlying data generation. For instance, in the case of electromagnetic brain signals, a forward model refers to the ``image'' that a current source in the brain \edit{generates at the scalp sensors}.  

For CorrCA, the backward model is given by the matrix ${\bf V}$ which maps observations ${\bf x}$ to components ${\bf y}={\bf V}^\top {\bf x}$. A corresponding forward model can be defined as the projection ${\bf A}$ that best recovers measurements ${\bf x}$ from the components ${\bf y}$ \citep{parra2002linear,parra2005recipes}: 
\begin{equation}
\hat{\bf x} = {\bf A} {\bf y}.
\end{equation}
The least-squares estimate of this forward model projection is
\begin{equation}
\hat{\bf A}={\bf R}_W {\bf V}({\bf V}^\top {\bf R}_W {\bf V})^{-1}.
\label{forward-model}
\end{equation}
Note that the columns ${\bf a}_k$ of this matrix capture how correlated a putative source ${\bf y}_k$ is with the observed sensor recordings ${\bf x}$. This approach of recovering the forward model from the backward model is explained in more detail in \cite{haufe2014interpretation}, along with a discussion on how forward and backward model parameters need to be interpreted. \edit{In the illustrative example of Figure~\ref{fig:toy-example-2D}, the forward model is shown in panel C as red and blue arrows for first and second component respectively. \edit{Evidently, the directions recovered by CorrCA match the underlying source orientations described in that example.}} 

\edit{The utility of the forward model will be demonstrated for the case of brain activity, where we are interested in how the activity of a recovered neural source manifests at the sensor level (Figure~\ref{fig:EEG-video-example} and Figure~\ref{fig:EEG-ERP}). These forward models can be used directly to perform source localization to identify the spatial origin of the corresponding current sources \citep{haufe2014interpretation}. In the case of questionnaire ratings (section~\ref{app:HBN-ratings}) we will interpret the meaning of a component by inspecting the forward model, as it measures the correlation of a component with answers of raters on different topics.}  
For the case of ratings, we \edit{may also be} interested in which items lead to a reliable aggregate score (sections~\ref{app:ratings}). \edit{In these cases, an analysis of backward model weights may provide valuable insight on what data features are most important to construct a reliable aggregate measure.}

Note that if the original data is uncorrelated, then ${\bf R}_W$ is diagonal and the projections ${\bf V}$ are orthogonal. In that case the forward and backward models are identical, except of an overall scale for each component. \edit{In that case, we can directly inspect the backward model, as is customarily done in PCA, which has orthogonal projections and one inspects the component ``loadings''. In the example of Figure~\ref{fig:HBN-ratings}D the covariance ${\bf R}_W$ is approximately diagonal and one can inspect the results with either the forward of backward model.}

\subsection{Relationship to multi-set Canonical Correlation Analysis}
\label{sec:mcca}

Thus far we have assumed that all repetitions should receive a common projection vector. This is sensible in the case of multiple trials if we expect that the sources of interest behave similarly across repetitions. However, in the case of \edit{diverse} subjects or raters, it may be more appropriate to assume that \edit{each repetition should be uniquely weighted}. For instance, brain activity recorded at a given location may differ between subjects due to anatomical or functional differences. This brings us to the case treated by Canonical Correlation Analysis (CCA) \edit{for $M=2$, or more generally,} multi-set CCA (MCCA) \edit{for $M>2$}. There are multiple implementations of MCCA \citep{kettenring1971canonical,nielsen2002multiset}. We will derive one of these instantiations as a maximization of ISC as defined in Eq. (\ref{isc}).  \edit{Assign to every subject $l$ a unique projection vector ${\bf v}^l$:}
\begin{equation}
y^l_i = {\bf v}^{l\top} {\bf x}^l_i . 
\label{projection-subject-dependent}
\end{equation} 
By inserting this into definition (\ref{isc}), it follows readily that
\begin{equation}
\rho =  
\frac{
\sum_{l=1}^N \sum_{k=1, k \neq l}^N {\bf v}^{l\top} {\bf R}^{lk} {\bf v}^k
}
{
(N-1) \sum_{l=1}^N {\bf v}^{l\top} {\bf R}^{ll} {\bf v}^l
} ,
\end{equation}
where ${\bf R}^{lk}$ is the cross-covariance matrix between ${\bf x}_i^l$ and ${\bf x}_i^k$:
\begin{equation}
{\bf R}^{lk} 
=
\sum_{i=1}^T 
({\bf x}_i^l - {\bar {\bf x}}_*^l)({\bf x}_i^k - {\bar {\bf x}}_*^k) ^\top.
\end{equation}

The projection vectors ${\bf v}^l$ that maximize $\rho$ can be found as the solution of  $\partial \rho / \partial {\bf v}^{l\top} = 0$, which yields for each $l$ the following equation: 
\begin{equation}
\frac{1}{N-1} \sum_{k=1, k \neq l}^N {\bf R}^{lk} {\bf v}^k
=
\rho \, {\bf R}^{ll} {\bf v}^l   .
\end{equation}
This set of equations for all ${\bf v}^l$ can be written as a single equation via concatenation to a single $ND$-dimensional column vector ${\bf v} = \left[ {\bf v}^{1\top} \ldots {\bf v}^{N \top} \right]^\top$:
\begin{equation}
{\bf R} {\bf v}
=
\lambda \, {\bf D} {\bf v} ,
\label{mcca-solution}
\end{equation}
where, $\lambda=(N-1)\rho+1$, ${\bf R}$ is a matrix combining all ${\bf R}^{lk}$, and where the covariance matrices ${\bf R}^{ll}$  are arranged in a block-diagonal matrix ${\bf D}$:
\begin{equation}
{\bf R} = 
\left[
\begin{array}{cccc}
{\bf R}^{11} & {\bf R}^{12} & \ldots & {\bf R}^{1N} \\
{\bf R}^{21} & {\bf R}^{22} & \ldots & {\bf R}^{2N} \\
\vdots & \vdots & \ddots & \vdots\\
{\bf R}^{N1} & {\bf R}^{N2} &\cdots & {\bf R}^{NN} \\
\end{array}
\right], 
{\bf D} = 
\left[
\begin{array}{cccc}
{\bf R}^{11} & 0 & \ldots & 0 \\
0 & {\bf R}^{22} & \ldots & 0 \\
\vdots & \vdots & \ddots & \vdots\\
0 & 0 &\cdots & {\bf R}^{NN} \\
\end{array}
\right].
\label{mcca-block-matrizes}
\end{equation}
If ${\bf D}$ is invertible, the stationary points of the ISC are now the eigenvectors of ${\bf D}^{-1}{\bf R}$, or more generally, the generalized eigenvectors of Eq. (\ref{mcca-solution}). We can arrange all such eigenvectors as the columns of a single matrix ${\bf V} = [{\bf v}_1 \ldots {\bf v}_n \ldots  {\bf v}_D]$. The eigenvector with the largest eigenvalue $\lambda$ maximizes the ISC because $\rho$ increases linearly with $\lambda$. The subsequent eigenvectors maximize ISC subject to the constraint that the component signals are uncorrelated from the previous components (see Appendix~\ref{apdx:max-sum-isc}).  In other words, the solution ${\bf V}$ diagonalizes ${\bf D}$:
\begin{equation}
{\bf V}^T {\bf D} {\bf V} = {\boldsymbol \Lambda}.
\label{constraint}
\end{equation}  

Formulating the MCCA problem as an eigenvalue problem can be derived in multiple ways  \citep{asendorf2015informative}. In \citet{kettenring1971canonical} it is presented as the maximization of the eigenvalue of the correlation matrix between subjects (MAXVAR criterion), but it can also be derived by maximizing the sum of correlations between subjects (SUMCORR criterion) \citep{nielsen2002multiset}. Here we have derived MCCA as a maximization of ISC. In this sense CorrCA can be seen as a version of MCCA constrained to have identical  projection vectors for all $N$ data sets (repeats/subjects). MCCA has $ND$ parameters per component, whereas CorrCA has only $D$. When there is an abundance of data ($T \gg ND$) and there is reason to believe that the $N$ datasets should be weighted individually, then MCCA may be appropriate \edit{\citep[e.g.,][]{de2018multiway}}. However, when $N$ is large compared to the data size  ($N>T/D$) CorrCA may be preferable \edit{due to the reduced number of free parameters}. This will be demonstrated in section~\ref{app:EEG-video-example}. 

\edit{When there are only two data sets ($M=2$) then the present MCCA algorithm reduces to conventional CCA \citep[][see Table 4.1]{borga1998learning}. In the illustrative example of Figure~\ref{fig:toy-example-2D}, CorrCA by definition gives the same result for both subjects (red and blue arrows), but we find that the directions derived with CCA differ between subjects (not shown). In section~\ref{app:HBN-ratings}, we will apply CCA to ratings data from two distinct raters (pairs of parent and child), in which case distinct projections may be warranted.}

\section{Applications}
\label{sec:apps}
In the following sections we demonstrate existing and new applications of CorrCA, while also examining issues related to statistical significance testing and regularization of the model. We also compare CorrCA with MCCA\edit{/CCA} and kernel-CorrCA. 

\subsection{Extracting brain activity that correlates between subjects}
\label{app:EEG-video-example}

We have previously used CorrCA to analyze brain activity recorded from multiple subjects  \citep{dmochowski2012correlated,dmochowski2014audience, ki2016attention,cohen2016memorable,poulsen2017eeg,cohen2017engaging,petroni2018variability,iotzov2017common}.  Briefly, subjects in these experiments watched movies or listened to narrated stories while their neural responses were recorded with multiple sensors (electroencephalography (EEG) --- electrical potentials measured at multiple locations on the scalp surface).  CorrCA was then used to extract components that were most reliable across subjects. These experiments showed that brain signals are significantly correlated between subjects on a rapid time-scale of less than a second (signals were high-pass filtered at 0.5~Hz). While the ISC values are small ($\rho \approx 0.05$), these values are statistically significant and very reliable across sessions \citep{dmochowski2012correlated} even in realistic environments \citep{poulsen2017eeg}. Interestingly, the level of correlation between subjects as measured by ISC is indicative of how attentive subjects are towards the stimulus \citep{ki2016attention} and, therefore, predictive of how well individuals remember the content of the videos \citep[see Appendix~\ref{apdx:individual-isc} for a definition of ISC for individual subjects]{cohen2016memorable}.  ISC of the EEG also predicts the retention of TV and online audiences \citep{dmochowski2014audience,cohen2017engaging}.

\begin{figure}[htbp]
\begin{center}
\includegraphics[width=\columnwidth]{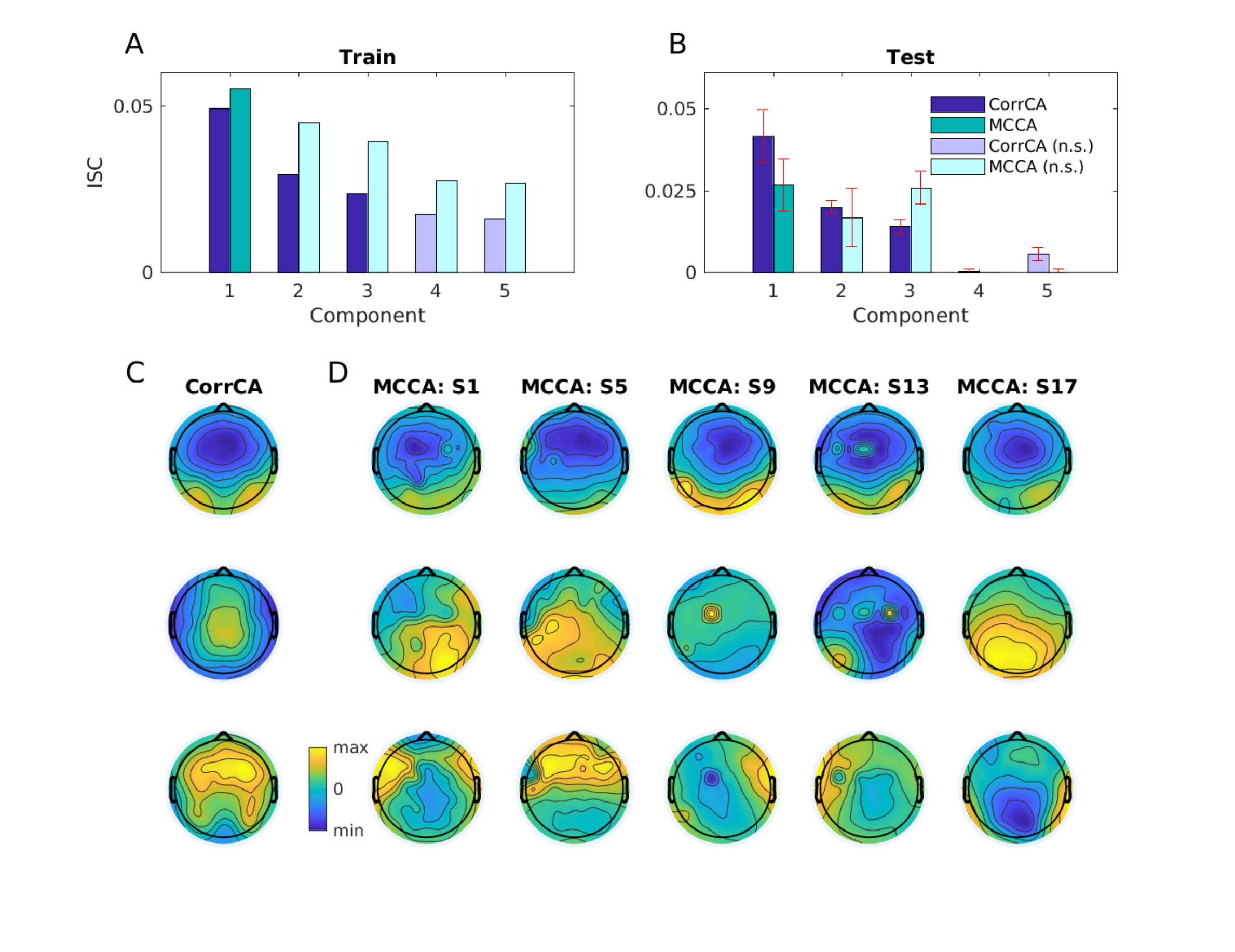}
\end{center}
\caption{
{\bf Correlated components of the EEG evoked by natural video stimuli.}  (A) The ISC achieved by both CorrCA (blue bars) and MCCA (green bars) on EEG data from 18 subjects viewing a 197-second video clip (``Sunday at Rocco's'', 2007, Storycorp). \edit{Here MCCA was regularized with TSVD ($K=12$; see Appendix~\ref{sec:regularization}), while CCA was not regularized. } Darker shades indicate that the learned components were found to be statistically significant using circular shuffle statistics at $\alpha<0.05$ (see Appendix~\ref{sec:significance}):  3 significant components were found with CorrCA, while 1 was identified with MCCA. Light shades indicate non-significant ISC. (B) ISC measured on independent \edit{test data}. \edit{Test-set performance was computed by training on 4/5 of the time samples and testing on the 1/4 of left-out samples.}. Here CorrCA produces higher ISCs than MCCA across 4 of the top 5 components (only the first is significant)  (C) The spatial distributions of the first three correlated components displayed here as ``forward models''  (see section~\ref{sec:forward-model}). The three components exhibit smooth topographies with unique spatial profiles, suggesting that each component recovers genuine and distinct neural activity. (D) The forward models of the subject-specific components learned by MCCA.   Each column displays the forward models of the components from an individual subject (i.e., subjects 1, 5, 9, 13, and 17).  The components are markedly less smooth, consistent with over-fitting, as MCCA learns separate parameters for each of 18 subjects.  For details on this data set, please refer to \citet{cohen2016memorable}.}
\label{fig:EEG-video-example} 
\end{figure}

Here we reanalyzed the data of \cite{cohen2016memorable} to compare CorrCA with MCCA (see section~\ref{sec:mcca}) on their ability to learn reliable projections of the data, while also using the opportunity to demonstrate forward model computation (section~\ref{sec:forward-model}) and statistical testing (Appendix~\ref{sec:significance}).   The data set consisted of the EEG from $N=18$ subjects as their brain activity was captured with $D=64$ scalp electrodes across $T=50344$ time samples corresponding to the 197-second video stimulus.  We first compared CorrCA with \edit{regularized} MCCA in terms of the number of reliable dimensions extracted by each, computed here using a non-parametric test using circular shuffle statistics (see Appendix~\ref{sec:significance}).   The ISC values found on the entire data set are shown for the first 5 components in Figure~\ref{fig:EEG-video-example}~A , where blue bars correspond to CorrCA and green to MCCA.  Moreover, darker shading indicates that the ISC of the component was found to be statistically significant: 3 significant dimensions were found by CorrCA, while \edit{only 1} was found by MCCA, despite the (training set) ISC values being higher for the latter.   
Indeed, when evaluating ISC on unseen \edit{test data the ISC drops substantially for MCCA as compared to the training data, and} CorrCA yielded higher ISC values than MCCA for 4 of the first 5 dimensions (Figure~\ref{fig:EEG-video-example}~B). 

\edit{Both of these findings are indicative of \edit{over-fitting with MCCA}, despite the use of regularization. We used the Truncated Singular Value Decomposition (TSVD, similar to what is described in Appendix~\ref{sec:regularization}) and selected a subspace of $K=12$ dimensions as this gave the largest test-set ISC (summed over the first 3 components). Note that MCCA fits a model for each subject. In this example it uses 216 parameters per component ($N=18$ subjects, $K=12$ dimensions), whereas CorrCA uses $D=64$ parameters per component on these data. While stronger regularization of MCCA is possible, the regularization is agnostic to the shared dimensions, which explains why ISC dropped with stronger regularization on the test data.} 
CorrCA can be seen as a form of regularization that addresses over-fitting\edit{, without removing dimensions that significantly correlate between subjects.} 

Both CorrCA and MCCA learned spatial filters that were applied to the multi-electrode EEG data  to maximize the ISC.  The activity captured by each spatial filter (component) is best depicted by ``forward models'' which are illustrated row-wise for the first three correlated components in Figure~\ref{fig:EEG-video-example}~C \& D.  The components extracted with CorrCA are marked by a smooth topography consistent with activation of the cerebral cortex (Figure~\ref{fig:EEG-video-example}~C). The components produced by MCCA are less smooth (Figure~\ref{fig:EEG-video-example}D shows subject-specific forward models for the first three components of 5 of the 18 subjects, i.e., subjects 1, 5, 9, 13, and 17).  Note that only the first component replicates, for some subjects, the bilateral occipital activation indicative of visual cortex activity seen in panel C (top). Other topographies are unlikely to be generated by a genuine neural source, and instead further support the assertion that MCCA overfit the data. 

\edit{Here and in the following examples we computed model parameters ${\bf v}$ by leaving out samples $i$, and measured test-set ISC on the left-out samples. The alternative could have been to leave out subjects. However, the number of entries in the between-subject correlation matrix ${\bf R}_B$ defined in Eq.~(\ref{between-subject-covariance-matrix}) scales quadratically with the number of subjects $N$, but linearly with number of samples $T$. Thus, we expect the accuracy of ${\bf R}_B$ and the resulting eigenvectors ${\bf v}$ in Eq.~(\ref{eigenvalue-equation-isc}) to  decrease faster when removing subjects instead of samples.}

\subsection{Maximizing inter-rater reliability}
\label{app:ratings}

\begin{figure}[htb]
\includegraphics[width=\columnwidth]{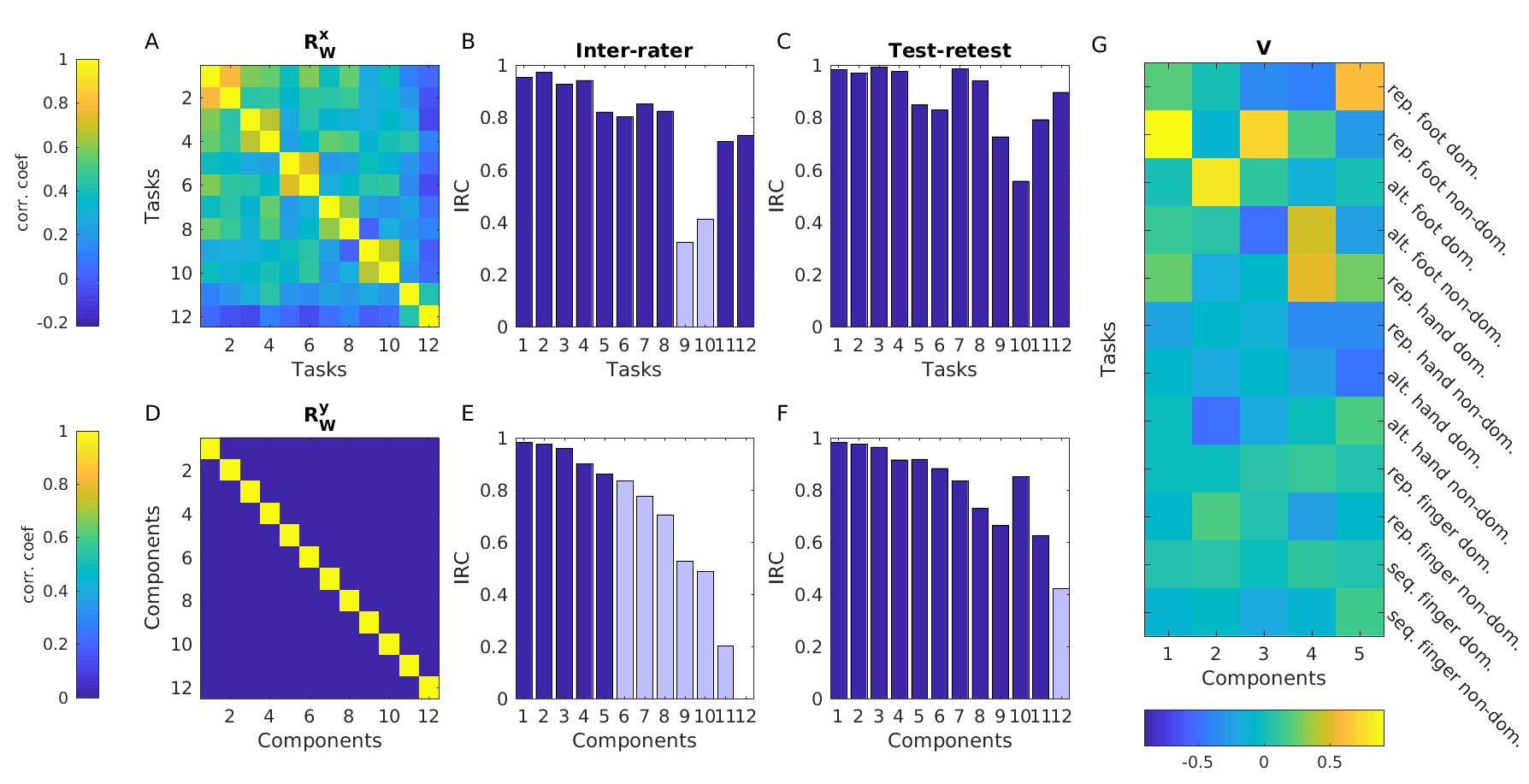}
\caption{{\bf Ratings from the Zurich Neuromotor Assessment}  for $T=30$ children, $D=12$ tasks that were rated, and $N=2$ clinical raters. The ratings are continuous numbers and normed by age. The goal of CorrCA is to identify aggregate scores that have high correlation between raters (IRC).  (A, B, C) Original rating data. (D, E, F) Components extracted by CorrCA. \edit{One of the two raters provided two repeated ratings for each child so that one can evaluate correlation between the repeated tests (test-retest). (A, E) show IRC between the two different raters. (B, F) show IRC between the two repeated ratings of the same rater.} Dark-shaded bars indicated statistically significant IRC (in panel E using circular shuffle statistics for training data; in panels B, C, F using F-statistics with Bonferroni correction; $\alpha<0.05$; see Appendix~\ref{sec:significance}) (G) Projection matrix ${\bf V}$ for the first 5 significant components. Data were provided by Valentin Rousson \citep{rousson2008reliability}. See text for details on each panel.}
\label{fig:ratings} 
\end{figure}

The theory laid out in section~\ref{sec:maximum-reliability} shows that one can use CorrCA to identify components with maximal inter-rater reliability. In this following example, clinicians assessed the motor skill of children by measuring performance on various standard tasks (Zurich Neuromotor Assessment). The rating obtained for each task here is the time taken by the child to perform each task. This time is normalized relative to the values of a standard group of children of the same age \citep{rousson2008reliability}. Although this is an objective measure, there is nonetheless variability due to variable performance of the child when repeating the task, and also subjective variability in the observer operating the stop-watch. In this specific data set, we have ratings from $T=30$ children on $D=12$ tasks from $N=2$ raters. The 12 measures are mostly positively correlated across the children (Figure~\ref{fig:ratings}~A, see panel G for descriptive task labels). Reliability between raters is measured here as the IRC, which differs across tasks (Figure~\ref{fig:ratings}~B). 

\edit{For many assessment instruments, individual ratings are summed up to obtain a total score. It is also natural to compute sub-scores by combining ratings on related items. In these sums, ratings effectively obtain equal weights, which is a sensible but nonetheless arbitrary choice. CorrCA can provide weightings that will give the most reliable response across raters, as reproducibility of results is a desirable goal.  In this specific example, the goal is to find the weightings of the 12 motor tasks that provide reliable aggregate (sub-) scores, and which capture independent aspects of these ratings (independent to second order).} The resulting weighting vectors ${\bf V}$ are shown in Figure \ref{fig:ratings}~G. Each column corresponds to a different component and has been normalized to have unit norm. 
Components are sorted in decreasing order of IRC (Figure \ref{fig:ratings}~E) and therefore decreasing order of inter-rater reliability (see section~\ref{sec:maximum-reliability}, Eq.~\ref{SNR}).  Of these components, the first five reach statistical significance at an $\alpha$-level of 0.05 (highlighted in Figure~\ref{fig:ratings}~E). Inspecting the ``backward model'' ${\bf V}$ it is clear that first component relies heavily on a single task (task 2, which has the highest IRC) but has small positive and negative contributions from other tasks. Components 2 and 3 are mostly a contrast between tasks 3 and 8, and 2 and 4, respectively. The detailed results can be obtained by running the code accompanying this article. In total, the method extracts different aggregate scores of high reliability, relying in some instances on a small subset of tasks. This has the potential of simplifying the assessment of motor performance while increasing its reliability.  An important aspect of these various components is that, by design, they are uncorrelated from one another (uncorrelated across the 30 children; Figure~\ref{fig:ratings}~D), whereas the original task ratings all measure similar aspects (they are predominantly positively correlated; Figure~\ref{fig:ratings}~A).

In this data set, children were also evaluated a second time by the same rater in order to assess test-retest reliability (Figure~\ref{fig:ratings}~C). Comparing panels B and C, it is evident that the test-retest \edit{reliability} is somewhat higher than the inter-rater \edit{reliability} (i.e., the evaluation does indeed depend to some degree on the subjective judgment of the rater). The components extracted with the highest IRC also exhibit the strongest test-retest correlations (at least for the first few components). This confirms that these aggregate measures generalize to data that was not used in the CorrCA optimization, and that minimizing subjective variability of the observer also minimizes test-retest variability.

\subsection{Identifying agreement between child and parent}
\label{app:HBN-ratings}

\begin{figure}[htb]
\includegraphics[width=\columnwidth,trim={5 0 50 10}]{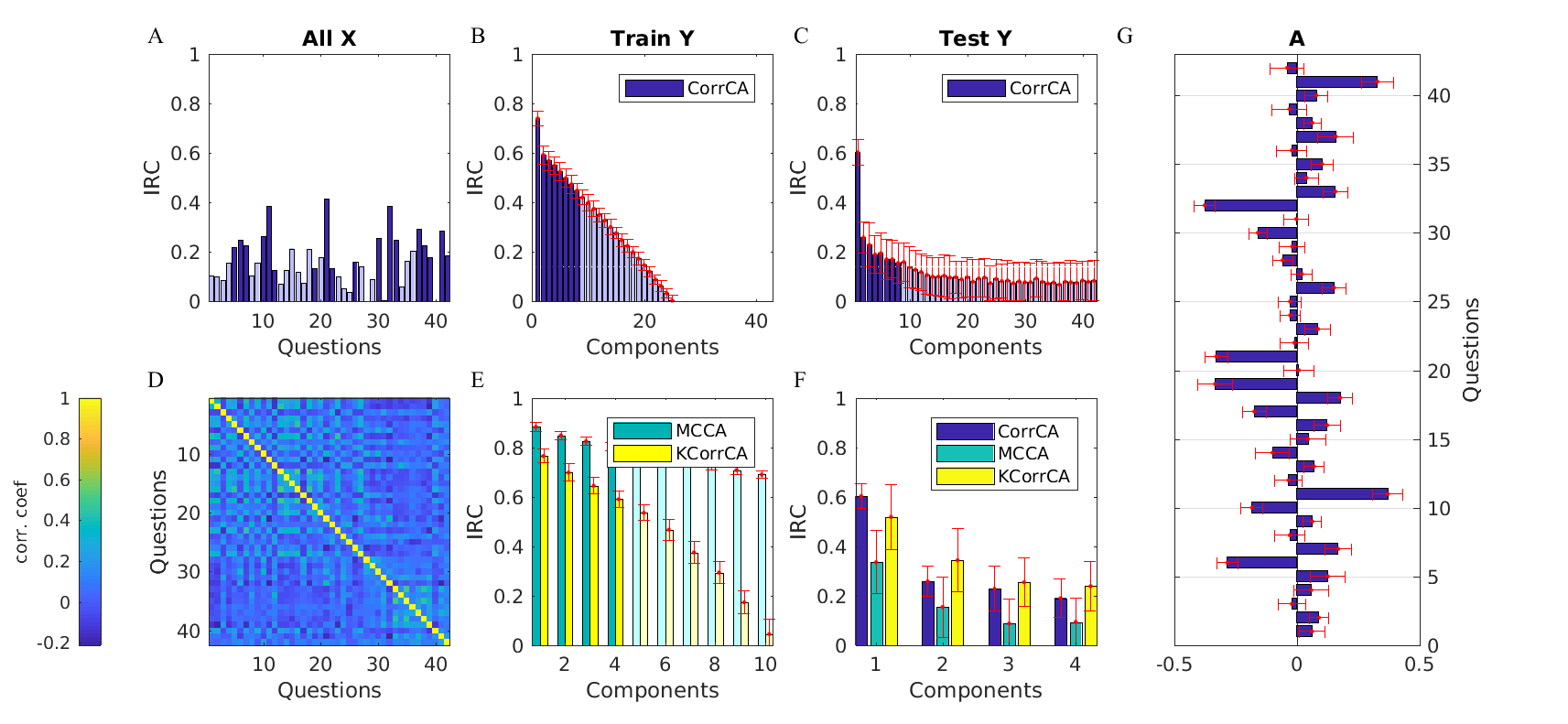} 
\caption{{\bf Ratings from the Alabama Parenting Questionnaire} for $T=616$ families and $D=42$ questions, with $N=2$ ratings from a child and one of its parents. Here, the goal of CorrCA is to maximize IRC, i.e., to identify agreement between child and parent. Data were collected by the Healthy Brain Networks Initiative \citep{Alexander149369}. Data were divided evenly and at random into a training and a test set. Second and third columns show results for training and test sets, respectively. Red error bars indicate the standard deviation across 100 random partitions of the data into training and test sets, and sampling with replacement within that (bootstrap estimates).  Dark-shaded bars indicate significant IRC at $\alpha<0.05$ (F-statistic,  Bonferroni corrected with $T=308$ for panels A and C; and circular shuffle statistic with $T=616$ for panels B and E).
(A) IRC between parent and child on all original questions. Dark-shaded bars indicate significant IRC (tested on Pearson's correlations, $\alpha=0.05, T=616$.   
(B) IRC for the components extracted with CorrCA on the training set. Dark-shaded bars indicate significant IRC (circular shuffle statistic, $\alpha<0.05, T=616$).
(C) IRC on the test set. Only the first component achieves strong IRC. In this case, we used shrinkage regularization ($\gamma=0.5$, see Appendix~\ref{sec:regularization}). Dark-shaded bars indicate significant IRC (F-statistic,  Bonferroni corrected, $\alpha<0.05, T=308$)
(D) Correlation between questions across all $616$ families. 
(E) IRC for components extracted with MCCA (section~\ref{sec:mcca}) and kernel-CorrCA (Appendix~\ref{sec:kernel-corrca}). Dark-shaded bars indicate significant IRC (circular shuffle statistic, $\alpha<0.05, T=308$)
(F) Comparison of IRC on the test set for the 4 strongest components extracted with CorrCA, MCCA and kernel-CorrCA. 
(G) Component weights for the strongest component extracted with CorrCA. } 
\label{fig:HBN-ratings}
\end{figure}

We also provide an additional example of CorrCA on subjective ratings data \edit{related to the style of parenting}. In this example, the goal is to identify questions on which parent and child agree. Specifically, we analyzed data collected with the Alabama Parenting Questionnaire \edit{\citep[APQ,][]{shelton1996assessment}} as part of the Healthy Brain Network initiative \citep{Alexander149369}. In the APQ, parents and children answered $D=42$ questions pertaining to their relationship by providing a numerical value from 1 to 5 for each question. From this, the APQ averages various questions to provide aggregate scores related to different aspects: ``involvement score'', ``positive parenting score'', ``poor supervision score'', ``inconsistent discipline score'', and ``corporal punishment score''.  At the time of this analysis, complete data were available for $T=616$ children/parent pairs (ages 6-17, 40\% female), i.e. $N=2$ raters. The first observation from the data is that the answers from children and parents are only weakly correlated ($\rho=0.17 \pm 0.1$, Figure~\ref{fig:HBN-ratings}~A). The aggregate scores fare even worse ($\rho=0.064 \pm 0.13$). A two-sample t-test suggests that these aggregate scores are less reliable than individual questions ($t(47)=2.4, p=0.022$) including several scores with negative correlations. Evidently, children do not much agree with their parents when it comes to parenting issues, despite answering \edit{nearly} identical questions on the topic.  

We applied CorrCA to find a combination of questions on which parents and children agree. To this end, we divided families at random into a training and a test set of equal size ($T=308$) and computed the projections ${\bf V}$ that maximized IRC on the training set. We then evaluated the resulting projections in terms of the correlation between child and parent for both train and test sets (Figure~\ref{fig:HBN-ratings}~B \& C). By design, IRC on the training set is largest for the first component. On the test set, the first component also has a relatively large IRC, whereas the other components do not differ substantially in IRC from the original set of questions (IRC hovers around 0.2 in the original ratings, ${\bf X}$, and the aggregate component scores, ${\bf Y}$). The lower IRC values on the test set as compared to the training set indicate the presence of over-fitting, although the discrepancy has been reduced here using shrinkage regularization (using $\gamma=0.5$; see Appendix~\ref{sec:regularization}). To determine how many significantly correlated dimensions are actually present in these data, we used circular shuffle statistics (see Appendix~\ref{sec:significance}) on all the data ($T=616$) and found $K=8$ components with $p<0.05$. On the test data alone, computation of the F-statistic revealed $K=8$ significant components (Figure~\ref{fig:HBN-ratings}~B \& C, dark shaded).

The component with the highest IRC has a combination of questions contributing to reasonably reliable aggregate scores ($\rho=0.61$ on test set, Figure~\ref{fig:HBN-ratings}~G).  \edit{To interpret this component we inspect its forward model weights (see section~\ref{sec:forward-model}.)}
The following \edit{six} questions correlate most strongly to the first component  (weightings of ${\bf V}$ are given in parenthesis): 
\edit{
Question 32: ``You are at home without an adult being with you'' (-0.38); Question 11: ``Your parent helps with your homework'' (+0.35); 
Question 19: ``Your child goes out with a set time to be home'' (-0.33); 
Question 21: ``You go out after dark without an adult with you'' (-0.33); Question 41: ``You use time-out as punishment'' (+0.33).
Question 6:  ``You fail to leave a note or let your parents know where you are going'' (-0.28). 
The sign of the forward weights indicate here whether the answer to that question correlate positively or negatively with the aggregate measure \citep{haufe2014interpretation}.} Therefore, this measure seems to capture the independence of the child versus parental involvement. \edit{The positive correlation of Question 41 with this independence-vs-involvement dimension is interesting as it suggests that this form of punishment is associated with a child's independence.} In total, while parents and children largely disagreed, there is moderate agreement on how independent their lives are. Given this result, we suspected that this metric increases with the child's age. Indeed, there is a strong correlation of this component with the child's age across the cohort (Pearson's $r=0.7$ and $r=0.67$ for parent and child scores respectively, $p<10^{-80}, T=616$; with no significant difference with the child's sex).  \edit{We did not analyze subsequent components due to the proximity of residual eigenvalues (Figure~\ref{fig:HBN-ratings}, panel C) and noise level may not permit accurate identification (see section~\ref{sec:identification-snr}).} 

Given that parents and children may genuinely differ in their judgments, it would make sense to allow for different weighting to the questions.  Therefore, we also applied MCCA to this data as described in section~\ref{sec:mcca}. We found only $K=4$ significant components at $p<0.05$ (as opposed to $K=8$ for CorrCA; Figure~\ref{fig:HBN-ratings}E, dark shaded). ISC summed over these first four components was numerically smaller for MCCA vs CorrCA (0.69 vs 1.10, values obtained on the test data without regularization; Figure~\ref{fig:HBN-ratings}~F). Therefore, it appears that on this data the sample size of $T=616$ was not sufficiently large to support a doubling in the number of free parameters (from 42 for CorrCA to 84 for MCCA). 

We have also tested the non-linear kernel-CorrCA method (see Appendix~\ref{sec:kernel-corrca}) on this ratings dataset (Figure~\ref{fig:HBN-ratings}~E). There was a small gain in terms of inter-rater correlation or inter-trial correlation respectively in the test data (Figure~\ref{fig:HBN-ratings}~F). However, the projections of the strongest components were not substantially different from the linear projections, suggesting that there were no strong non-linear relationships in the data. We find a similar result with kernel-CorrCA for the the data presented in the next section \ref{app:EEG-ERP} (results not shown).   

\subsection{Identifying stimulus-related components in EEG with high SNR}
\label{app:EEG-ERP}

PCA is often used to extract a low-dimensional representation of high-dimensional data. To this end, PCA selects orthogonal projections that capture most of the variance of the projected samples. When multiple repetitions of the data are available, CorrCA can be used for the same purpose of reducing dimensionality of the data, but with the objective of capturing dimensions that are reliably reproduced across repetitions. Here we apply this to a conventional scenario in neuroimaging: neural activity is collected over time with multiple sensors, while the same experimental stimulus is repeated in multiple trials. We propose to use CorrCA to identify spatio-temporal patterns of activity that are well preserved across trials, similar to what was proposed in \citep{tanaka2013task,de2014joint,dmochowski2015maximally,dmochowski2015cortical}. Ideally, the corresponding time-courses of multiple components are uncorrelated in time so that they capture different activity. CorrCA guarantees this by design as it diagonalizes ${\bf R}_W$. CorrCA is therefore similar to other source separation methods such as Independent Components Analysis \citep{makeig1996independent}, Denoising Source Separation \citep{sarela2005denoising}, or Joint-Decorrelation  \citep{de2014joint}. In fact, CorrCA can be viewed as a form of blind-source separation, as no labels are required for the analysis.

\begin{figure}[htbp]
\includegraphics[width=\columnwidth,trim={80 50 100 50}]{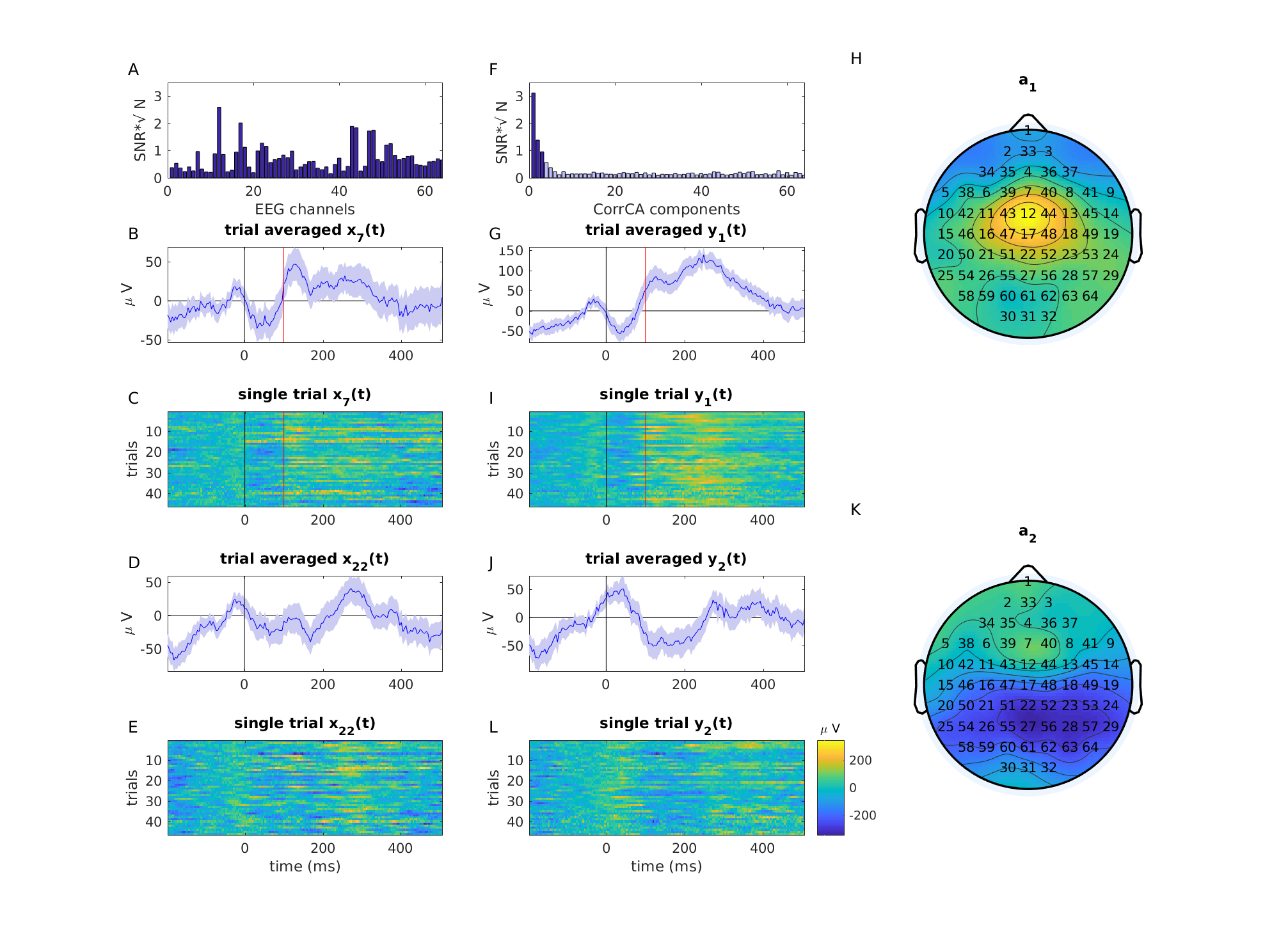}
\caption{{\bf Event-related potentials.}  EEG recordings from a single participant collected during a speed two-alternative forced-choice task (Eriksen Flanker Task). (A-E) Original EEG data $x_j(t)$, highlighting two channels, $j=7$ (Fz electrode, SNR=0.97) where the ``error-related negativity'' is expected to be most pronounced (between 0 and 100ms; black and red vertical lines), and $j=22$ (PCz electrode, SNR=1.3) where a ``readiness potential'' is typically observed.  (F-L)  First and second correlated components with the strongest inter-trial correlation. (A \& F) SNR computed from the inter-trial correlation (Eq.~\ref{SNR}) for original data and extracted components respectively. Dark shaded bars indicate significant components (Circular shuffle statistic on the training set; $\alpha<0.05$; see Appendix~\ref{sec:significance}) (B, D, G, F) Average across $N=46$ repeated trials. Blue-shaded area indicates standard-error of the mean. Evidently the extracted components have a smaller SEM relative to the mean values. The SNR measures are multiplied with $\sqrt{N}$ to parallel the SEM shown here. SNR of 2, for instance, indicates that the mean is 2 SEM away from zero baseline. (C, I, E, F) EEG resolved over all trials to visualize the inter-trial variability. (H \& K) Forward model of the two strongest components with distributions consistent with the ERN and readiness potential respectively. Data from \citep{parra2002linear}.} 
\label{fig:EEG-ERP} 
\end{figure}

Here we analyze EEG data that were collected during an Eriksen Flanker Task \citep{parra2002linear}. Subjects were presented with a collection of arrows pointing left or right. They were asked to decide as quickly as possible between two alternative choices (the direction of the center arrow) by pushing corresponding buttons with either their left or right hand. We analyze data from a single participant who responded correctly in 550 trials, and made an error in 46 trials (Figure~\ref{fig:EEG-ERP}). This error, which participants notice as soon as they make their incorrect response, leads to a negative EEG potential 0-100~ms after the button press (this period is indicated between black and red vertical lines in panel~B \& C for electrode \#7, see electrode location in panel~H). This phenomenon is known as the ``error-related negativity'' \citep[ERN,][]{yeung2004neural}.   
Button pushes also elicit the ``readiness potential'' --- a gradually increasing potential preceding the button push. Aside of the readiness potential and error-related negativity there are a series of other potential deflections following the button push. In this example, we have $T=176$ time samples, $D=64$ electrodes, and $N=46$ trials. Time is aligned across trials \edit{so that} the button press is at time $t=0$ ms. The traditional approach to analyzing such data is to average the electric potentials over trials, yielding what are called event-related potentials (ERPs, panel~B). The reliability of these trial-averages is relatively low (panels~B\& D show mean and standard-error \edit{across trials}, and panels~C~\& E show single trial data). We selected  electrodes \#7 and \#22 for display as these capture distinct time courses and locations identified in prior literature. The SNR (as defined in section \ref{sec:maximum-reliability}) is maximal for the same cluster of frontal electrodes (\#12, \#17, \#44, \#43) and the mean SNR across all electrodes is $S\sqrt{N} = 0.72$ (Figure~\ref{fig:EEG-ERP}~A). We are multiplying \edit{SNR} with $\sqrt{N}$ because, typically, the goal is to determine if the mean value differs significantly from zero, and thus standard-error of the mean \edit{(SEM)} is the relevant quantity.

\edit{In our earlier work we specifically looked for a projection that distinguishes the EEG evoked response between error and correct responses \citep{parra2002linear}. Here the goal is to identify reliable error responses.} When we apply CorrCA to these data, we obtain a series of components with high inter-trial correlation (Figure~\ref{fig:EEG-ERP}~F). CorrCA components were extracted using shrinkage regularization ($\gamma=0.4$, see Appendix~\ref{sec:regularization}) on a training set of $N=45$ trials (leaving 1 trial out for testing). Performance was evaluated on the leave-one-out test set. The first three components (highlighted in dark blue) are significant at the $\alpha=0.05$ level according to a non-parametric test based on circularly shifted surrogate data (see Appendix~\ref{sec:significance}). The first two of these components have an SNR of $S\sqrt{N} = 3.3$ and $1.6$. Evidently, component 1 is very reliable (panel~G shows that the mean is several SEM away from the baseline). The deflections are now obvious even in single trials without the need to average (panel~I). The spatial distribution of this component activity (panel~H) is indicative of the well-known ERN. Component 2 (panel~J) is indicative of the readiness potential \citep{deecke1969distribution}. It rises leading up to the button push at time $t=0$ ms, and has a spatial distribution over parietal and motor areas (panel~K).

Note that CorrCA has reduced the $D=64$ dimensional data to just a few components that capture the reliable portion of the data (high SNR; panel~F).  By compactly representing neural activity in a small number of components, it becomes easier to probe for changes in activity due to manipulation of experimental variables, whose effects on the original $D=64$ dimensional data may be obscured due to the lower SNR and the multiple comparison problem.  

\edit{Like most data decomposition algorithms, CorrCA does not require that the underlying data have any time structure. For time series data, this implies that the correlated signal of interest is observed with the same delay in each channel and repetition. If this assumption is violated, drops in performance will occur.}

\subsection{Effect of regularization on the reliability of EEG responses}
\label{app:regularization-evaluation}

A common numerical difficulty with generalized eigenvalue problems is the required matrix inverse. In the case of CorrCA, it is necessary to invert ${\bf R}_W$ in equation~(\ref{eigenvalue-equation-isc}). When some of the eigenvalues are small (or even zero), then random fluctuations due to noise will dominate the inverse. It is thus important to regularize the estimate of ${\bf R}_W$. In the past, we have used the Truncated Singular Value Decomposition (TSVD) \citep{hansen1994regularization} and shrinkage \citep{ledoit2004well} for this purpose \citep{ dmochowski2012correlated,dmochowski2014audience,ki2016attention,cohen2016memorable}.  These methods are described in detail in Appendix \ref{sec:regularization}. 

To evaluate the effects of these regularization methods on the results of CorrCA we use data from our study on neural responses to advertisements \citep[we select one commercial among several aired during the 2012 Super Bowl]{dmochowski2014audience}.  For both the TSVD and shrinkage approaches, we spanned the space of regularization strength (a single parameter, $K$ and $\gamma$, in each case) and measured ISC.  The training set was the EEG data from 12 subjects during the first viewing of the ad, while the EEG data collected during the second viewing served as the test set.  

\begin{figure}[htbp]
\includegraphics[width=\columnwidth,trim={60 30 50 10}]{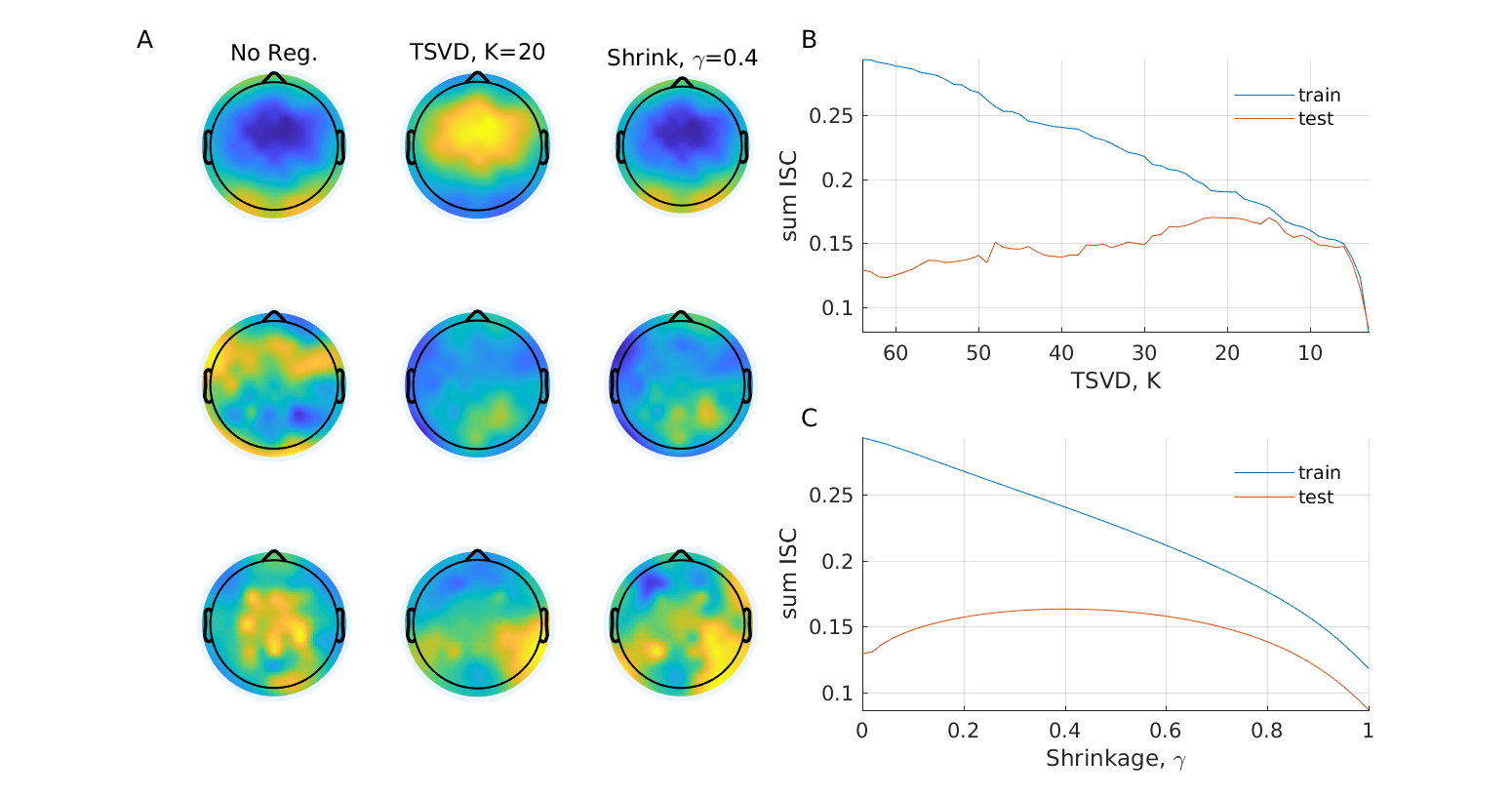}
\caption{{\bf Regularization increases reliability of EEG test data.}  The inter-subject correlations of EEG responses from 12 subjects viewing a popular television advertisement \citep{dmochowski2014audience}.  We took the data of the highest-scoring advertisement maximized for ISC while varying levels of TSVD and shrinkage regularization. {(A)}  The spatial distributions of the three most correlated components.  Note that the distributions of the second and third components were distinct for unregularized, TSVD-regularized and shrinkage-regularized CorrCA.  {(B)} Performance with TSVD as a function of $K$ (decreasing $K$ corresponds to stronger regularization), and {(C)} performance with shrinkage as a function of $\gamma$ (increasing $\gamma$ corresponds to stronger regularization). Components were extracted on the data from a first viewing of the advertisement (training) and evaluated on data from a second viewing (test).  Reliability peaks at $\gamma=0.4$ for shrinkage, and when truncating ${\bf R}_W$ to $K=20$ dimensions, with very similar peak values of $\approx 0.16$ (sum of first three inter-subject correlations) for the two regularization approaches.}
\label{fig:regularization} 
\end{figure}

The sum of the first 3 ISC values for both regularization methods followed a similar trajectory with increasing regularization strength (decreasing $K$ and increasing $\gamma$; Figure~\ref{fig:regularization}~B \& C). On the training data, ISC drops monotonically, whereas on the test data a maximum is achieved at intermediate regularization values. On these data, shrinkage regularization was most effective at $\gamma=0.4$ and TSVD when retaining $K=20$ out of a total of $D=64$ dimensions in the within-subject covariance ${\bf R}_{W}$.  The forward models of the first three correlated components for each regularization technique are shown in Figure~\ref{fig:regularization}~A.  Notice that regularization tends to smooth out the topographies of the second and third components compared to no regularization.    

\section{Validation on simulated data}
\label{sec:simulations}

We evaluated the conditions under which CorrCA identifies the true source directions. We also determined the performance of CorrCA at identifying components with high ISC, and compared this against a simple benchmark based on PCA of the subject-average data. Additionally, we compared the accuracy of the statistical tests described in Appendix~\ref{sec:significance} at estimating the correct number of components. To this end, we generated artificial data with known properties. The data are generated assuming a known number of correlated components (signals \edit{shared across subjects}) and additive spatially correlated noise \edit{not shared across subjects}. The following factors were varied parametrically: the number of subjects (raters, repeats), $N$, the number of samples per subject, $T$, the number of dimensions, $D$, the number of correlated components, $K$, and the signal-to-noise ratio (SNR) as defined in the space of multivariate measurements. Each simulated correlated component was identically reproduced in each subject, so that the ISC of all components was equal to one. We further varied the temporal dependency structure of the signal and noise components (IID or pink noise), the distribution of the signal and noise components (Gaussian or $\chi^2$ distributed), and the heterogeneity of the signal and noise subspaces (either the same or different for all subjects). Signal components were generated to be correlated across subjects, but uncorrelated with each other. Noise components were uncorrelated with each other as well as across subjects. The number of noise components was fixed to $D$. The following default parameters were used unless otherwise noted: $N=5$, $T=200$, $D=30$, $K=10$, SNR~=~0\,dB, no regularization, IID Gaussian signal and noise components, ISC = 1 for all $K$ components. Details on data generation and evaluation methods can be found in Appendix~\ref{sec:simulation-evaluation-methods}.

\subsection{Accuracy as a function of Signal-to-Noise Ratio}
\label{sec:identification-snr}

\begin{figure}[htbp]
\centering
%\bfseries \sffamily A 
%\begin{tabular}{cc}
%\includegraphics[width=0.45\columnwidth]{sim2a_ISC_pca_mean.png} &
%\includegraphics[width=0.45\columnwidth]{sim2a_recon.png}\\
%\includegraphics[width=0.45\columnwidth]{sim2a_K.png} &
%\includegraphics[width=0.45\columnwidth]{sim2b_K.png}\\
%\end{tabular}
\includegraphics[width=0.9\columnwidth]{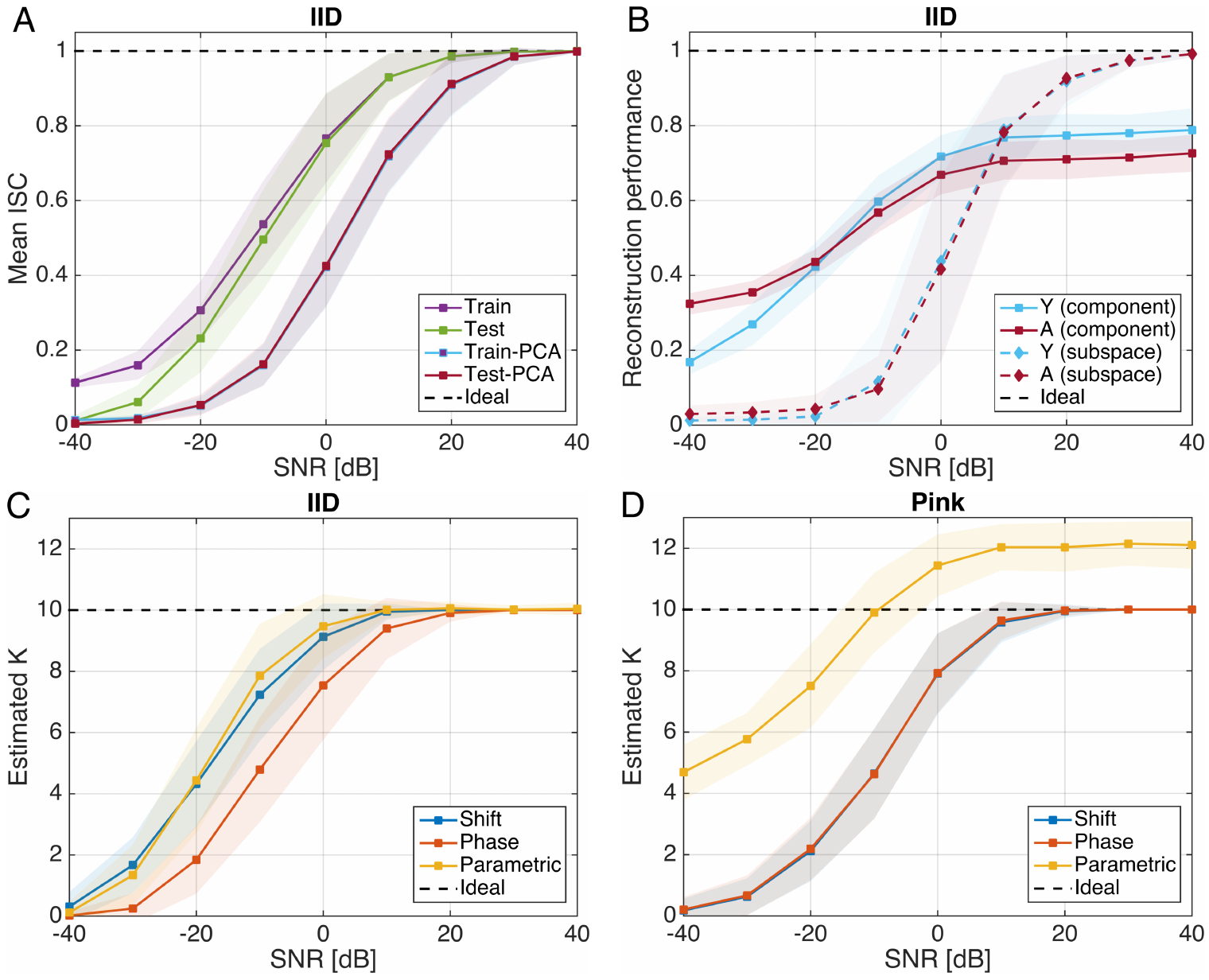}
\caption{{\bf CorrCA performance as a function of signal-to-noise ratio (SNR).} Other parameters were fixed to $T=200$, $D=30$, $N=5$, $K=10$. Plotted are averages across 100 experimental repetitions, while corresponding standard deviations are shown as transparent areas. (A) The average ISC of the reconstructed components converges to the maximal value of 1 for very high SNR. For very low SNR, the expected value of 0 is exceeded on training data, indicating overfitting. CorrCA outperforms a simple PCA on the subject-averaged data. (B) CorrCA finds to correct subspace of correlated components if the SNR is sufficiently high. Individual components may, however, not be identifiable if their ISC are too similar. (C)  In case of IID data, all statistical tests to estimate $K$ converge to the true number of 10 for high SNR. (D) In case of dependent (pink noise) samples, only surrogate data approaches perform well, while the parametric test overestimates $K$.}
\label{fig:sim-SNR-dependence}
\end{figure}

Figure~\ref{fig:sim-SNR-dependence} depicts CorrCA performance as a function of measurement SNR for fixed $T=200, D=30, N=5, K=10$. SNR values range from -40\,dB to 40\,dB. It is apparent that average training and test ISC reach the optimal value of 1 for very high SNR (panel~A). For very low SNR ranges, training ISC significantly exceeds test ISC, indicating over-fitting. The identification of the correlated subspace converges to the optimal value for high SNR (panel~B). However, individual components are not perfectly identifiable even for high SNR because they share the same ISC values (see Appendix~\ref{apdx:identifiability} for a discussion). The identifiability of individual components becomes possible if the ISC of the simulated signal components are adjusted to linearly decrease from 1 down to 1/$K$, and if the measurement SNR is high (data not shown). In practice, one should keep in mind that the uniqueness of the ISC (and thereby, their individual identifiability) depends on the amount of noise that is collinear to each original correlated dimension. In the IID case, all estimates of $K$ converge to the true number of 10 for high SNR (panel~C). Here, the parametric test and the surrogate data approach using random circular shifts perform similarly, while the surrogate data approach using phase scrambling performs considerably worse at low SNR, presumably due to the absence of a well-defined notion of frequency and phase in the sense of the Fourier transform. In the case of dependent (pink noise) samples, both surrogate data approaches perform identically, converging to the true value of $K$ for high SNR (panel~D). In contrast, the parametric test, which incorrectly assumes IID data, overestimates the number of correlated components across the entire SNR range, as expected. 

Note that the significance test is applied here to multiple components ($D=30$). Under appropriate conditions, all three methods converge to the correct number of dimensions in these simulate data ($K=10$). This indicates that all three methods (described in Appendix~\ref{sec:significance}) properly account for these multiple comparisons.

\subsection{Accuracy of estimating dimensionality of correlated components}

\begin{figure}[htbp]
\centering
\includegraphics[width=0.9\columnwidth]{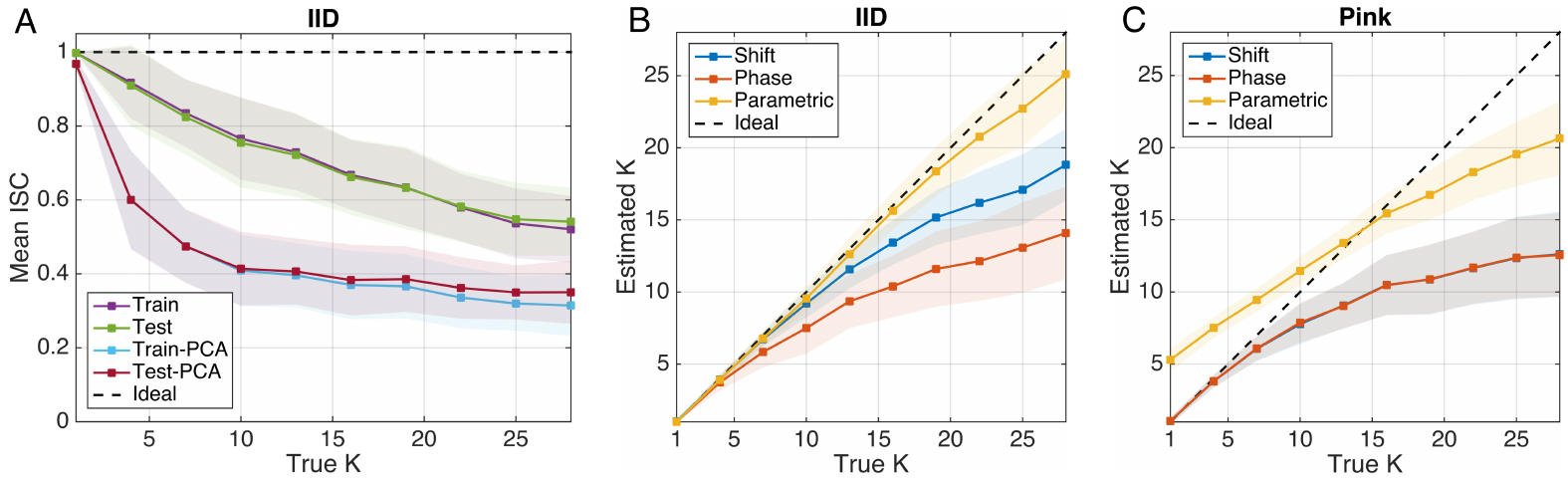}
\caption{
{\bf CorrCA performance as a function of the true number of correlated components, K.} Other parameters were fixed to $T=200$, $D=30$, $N=5$, SNR~=~0\,dB. (A) The average ISC of the reconstructed correlated components drops with increasing $K$. (B) Consequently, the number of correlated components can be less well estimated, the higher $K$. For IID data, this number is bound from above by the true number for all statistical approaches. (C) Overestimation of $K$ may occur for pink noise data when the parametric F test is used. } 
\label{fig:sim-varying-K}
\end{figure}

The performance of CorrCA as a function of the true number of correlated components, $K$, is depicted in Figure~\ref{fig:sim-varying-K}. Here, $T=200$, $D=30$, $N=5$, SNR~=~0\,dB, where $K$ was varied from 3 to 28. As the overlap between signal and noise subspaces increases with $K$, the average ISC of the reconstructed correlated components drops (panel~A). A similar decline is seen in the reconstruction of simulated correlated components and their forward models (data not shown). As a result, the percentage of correlated components that can be recovered (i.e., reach statistical significance) also decreases with increasing $K$  (panels~B \& C).

We also compared the ISC values obtained with CorrCA against a simple benchmark, namely, PCA of the subject-averaged data. While this simple method captures the correct directions for high SNR case, CorrCA significantly outperforms at medium and low SNR values (Figure~\ref{fig:sim-SNR-dependence}~A). The same is true for simulations with varying number of underlying dimensions (Figure~\ref{fig:sim-varying-K}~A), where we see that CorrCA outperforms this simple method in all instances tested.

\begin{figure}[htbp]
\centering
\includegraphics[width=0.9\columnwidth]{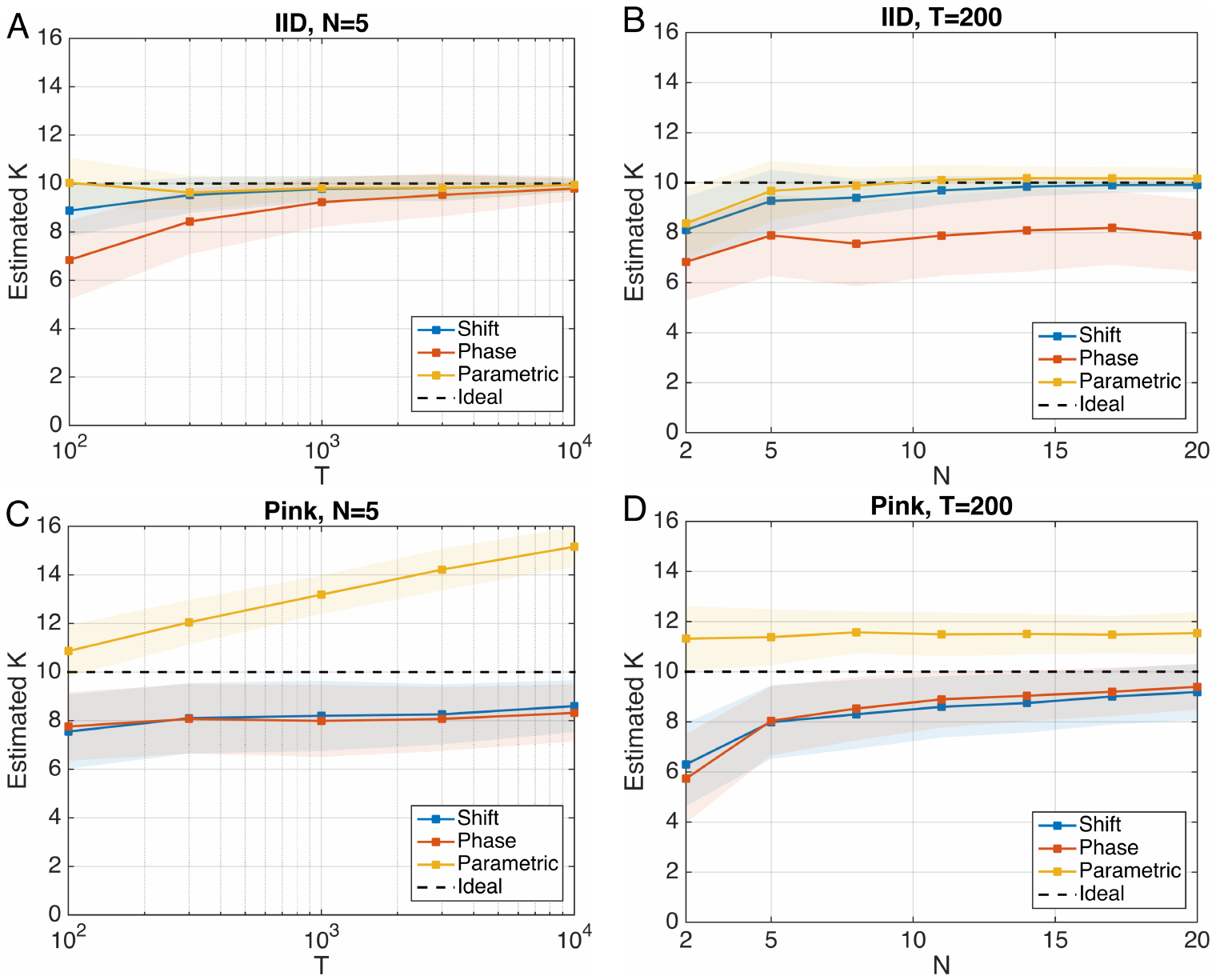}
\caption{
{Estimated number of correlated components as a function of the number of data points, T, (left panels), and the number of repetitions (subjects/raters), N, (right panels).} Other parameters were fixed to $D=30$, $K=10$, SNR~=~0\,dB. (A, B) In case of IID data, all estimates converge to the true value of 10 for increasing $T$ or $N$. (C, D) For pink noise data, only non-parametric statistical tests converge to the true value.}
\label{fig:sim3}
\end{figure}

Figure~\ref{fig:sim3} shows the estimated number of correlated components as a function of the number of data points, $T$, and the number of repetitions (subjects/raters), $N$, with fixed $D=30$, $K=10$, SNR~=~0\,dB. In case of IID data, all estimates converge to the true value of 10 for increasing $T$ or $N$. This is also true for pink noise data in combination with non-parametric statistical tests. For the parametric test, however, the number of seemingly correlated components diverges as $T$ increases, and is overestimated regardless of $N$.

\subsection{Dependence on signal and noise distributions}

\begin{figure}[htbp]
\centering
\includegraphics[width=0.9\columnwidth]{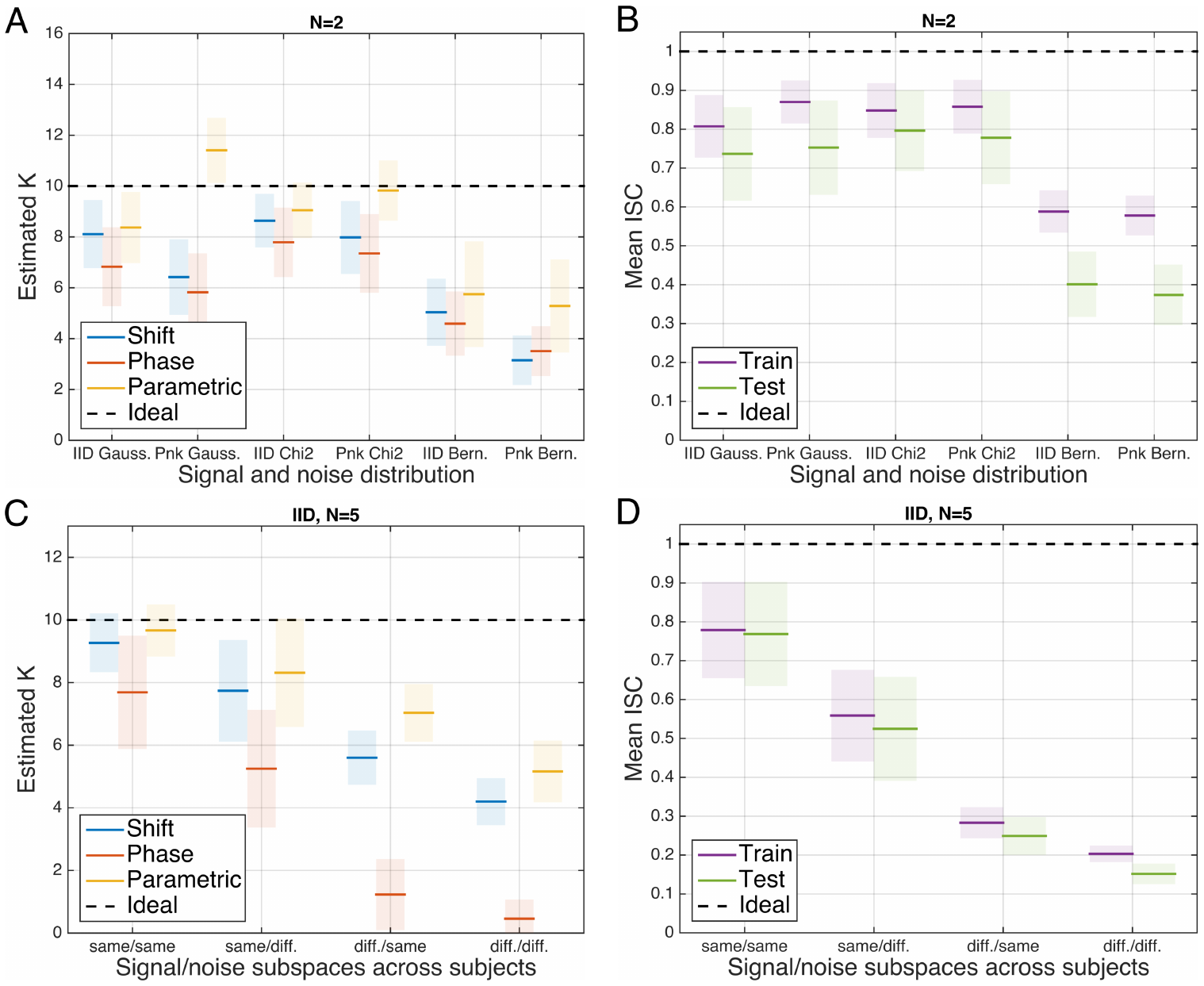}
\caption{
{\bf CorrCA performance as a function of signal and noise distributions.} Other parameters were fixed to $T=200$, $D=30$, $K=10$, SNR~=~0\,dB. (A, B) Performance for Gaussian and squared Gaussian ($\chi^2$-distributed) signal and noise components, as well as for dichotomized (Bernoulli-distributed) observations. Data points were either drawn IID or sampled from a pink noise process. While non-Gaussianity of the underlying components does not substantially affect the extraction of correlated components, the same is considerably more difficult in the presence of dichotomous observations. (C, D) performance on IID Gaussian data when signal and noise covariance structure differs across repetitions (e.g., subjects, raters). Significant drops in the number of estimated correlated components as well as the average ISC are observed if the signal covariance, the noise covariance, or both, differ across subjects. }
\label{fig:sim4}
\end{figure}

Figure~\ref{fig:sim4} A \& B shows the performance of CorrCA for different signal and noise distributions. Here, $T=200$, $D=30$, $N=2$, $K=10$, SNR~=~0\,dB. Signal and noise components were generated either as Gaussian or squared Gaussian ($\chi^2$-distributed). For Gaussian distributed components, we considered an additional case, where observations ${\bf x}^l(t)$ were dichotomized (thus, Bernoulli distributed) by taking their sign. This simulation scenario resembled the typical setting in which two raters provide a binary assessment of the same issue. While non-Gaussianity of the underlying components does not substantially affect the extraction of correlated components, the same is considerably more difficult in the presence of dichotomous observations. An additional analysis was conducted for IID Gaussian data and $N=5$ to assess the performance of CorrCA when signal and noise covariance structures (as defined by the mixing matrices ${\bf A}_\text{s}^l$ and ${\bf A}_\text{n}^l$, $n = 1, \hdots, N$) differ across subjects (Figure~\ref{fig:sim4} panel C \& D).  Significant drops in the number of estimated correlated components as well as the average ISC are observed if the signal \edit{directions}, the noise  \edit{directions}, or both, differ across subjects.  \edit{Note that varying directions of the non-reproducible component ${\bf A}_\text{n}^l$ across subjects is equivalent to adding noise, so that the dependence on the noise directions shown in Figure~\ref{fig:sim4} (panel C \& D) is a restatement of the dependence on SNR of Figure~\ref{fig:sim-SNR-dependence}.}  The result illustrates that identical \edit{directions} for the  signal components across subjects are a fundamental assumption of CorrCA. If this assumption is suspected to be violated, the use of MCCA (section~\ref{sec:mcca}) may be beneficial, or approaches that can gradually move from CorrCA to MCCA \citep{kamronn2015multiview}.

\section{Conclusion}

The goal of this work was to provide a formal, yet didactic and comprehensive treatment of this new component analysis technique. The analytic development resulted in a direct link to MCCA and a surprising equivalence to LDA, thus linking the new methods with classic concepts of multivariate analysis. We also found an efficient scheme for computing ISC/IRC which surprisingly does not require computation of pairwise correlations. We also identified  the F-statistic as an exact parametric test for statistical significance, which is valid in the case of normal and identically distributed samples. For the case of dependent samples we demonstrated and validated circular-shift shuffle statistics as an efficient non-parametric test of significance. 
We also presented an extension to non-linear mappings using kernels. 
Finally, perhaps the most important contribution of this work is the formalization of inter-subject correlation. Conventional metrics, such as pairwise Pearson's correlation are limited to two signals. Inter-subject correlation as defined here is applicable to more than two signals. ISC differs from conventional definitions of correlation (see section~ref{apdx:ICC}), yet seamlessly integrates in well established mathematics related to linear discriminants, canonical correlation, and the F-statistic. 

We make all code and data available at \url{http://parralab.org/corrca},  and for long-term storage here \cite{code_data_url}. This code is significantly faster than previous instantiations and has been extensively evaluated. All figures presented in this paper can be reproduced with this new code base and associated data. We hope that this analysis and code inspires new uses for correlated component analysis. 

\section{Acknowledgments}

We are grateful to Valentin Rousson who pointed us to work on inter-class correlation, and Oskar Jenni of the Children University Hospital in Zurich, Switzerland who collected  the data published in \citep{rousson2008reliability}, which we reanalyzed here in Figure~\ref{fig:ratings}. Similarly, we would like to thank Mike Milhalm for pointing us to the problem of identifying parent/child agreement and Lindsay Alexander for providing access to the Healthy Brain Network data \citep{Alexander149369} for Figure~\ref{fig:HBN-ratings}. We also thank Samantha Cohen for providing the data from \citep{cohen2016memorable} for Figure~\ref{fig:EEG-video-example}. We thanks Samantha Cohen, Jens Madsen Michael X Cohen, and two anonymous reviewers for useful comments on an earlier version of this manuscript. We want to thank Flaviano Morone for assisting with the proof in Appendix~\ref{apdx:max-sum-isc}. 

% \section*{Author contribution}
%Lucas Parra developed the math and wrote much of the code and text for this paper. Stefan Haufe provided an early implementation of the ISC maximization algorithm which motivated the definition used here for ``within'' and ``between'' subject covariance matrices. He also provided all the simulations and code related to shuffling statistics to establish significance and performance of the algorithms on simulated data and wrote sections \ref{sec:significance}, \ref{sec:simulations} and\ref{sec:simulations}. Jacek Dmochowski was the first to implement CorrCA for $N=2$ data sets and he extended it to $N>2$ by concatenating all possible pairs of subjects \citep{dmochowski2012correlated}, something that is close but not identical to the approach presented here. Jacek worked on much the the issues related to regularization and wrote most of section \ref{sec:regularization}. All authors carefully edited the entire manuscript. 

\appendix 

\section{Generalized eigenvectors for two symmetric matrices }

\subsection{Joint diagonalization}
\label{apdx:joint-diagonalization}

Here we will review the well-know relationship between the joint diagonalization of two symmetric matrices ${\bf A}$ and ${\bf B}$ and the eigenvectors of ${\bf B}^{-1}{\bf A}$. Consider eigenvectors arranged as columns in ${\bf V}$ with the corresponding eigenvalues in the diagonal matrix ${\boldsymbol \Lambda}$: 
\begin{equation}
{\bf A} {\bf V} = {\bf B} {\bf V} {\boldsymbol \Lambda} .
\label{generalized-eigenvalue-equation}
\end{equation}
To find the solution ${\bf V}$ for this equation, we can replace the positive definite matrix ${\bf B}$ with its Cholesky factorization
\begin{equation}
{\bf B}={\bf L} {\bf L}^\top. 
\label{cholesky-factorization}
\end{equation}
Insert this into (\ref{generalized-eigenvalue-equation}) and left-multiply with ${\bf L}^{-1}$ to obtain:
\begin{eqnarray}
({\bf L} ^{-1} {\bf A} {\bf L}^{-\top} ) ({\bf L}^\top {\bf V})& = &({\bf L}^\top {\bf V} ) {\boldsymbol \Lambda} , \\
{\bf M} {\bf U} &= & {\bf U}  {\boldsymbol \Lambda} ,
\label{eigenvalue-equation}
\end{eqnarray}
with ${\bf M} = {\bf L} ^{-1} {\bf A} {\bf L}^{-\top}$ and ${\bf U} = {\bf L}^\top {\bf V} $. Equation (\ref{eigenvalue-equation}) is a conventional eigenvalue equation with the same eigenvalues as (\ref{generalized-eigenvalue-equation}). It is now easy to see that the corresponding eigenvectors ${\bf V} = {\bf L}^{-\top} {\bf U}$ diagonalize matrix ${\bf A}$
\begin{equation}
{\bf V}^\top {\bf A}{\bf V} = {\bf U} {\bf L}^{-1} {\bf A} {\bf L}^{-\top} {\bf U}= {\bf U}^\top {\bf M} {\bf U} = {\boldsymbol \Lambda} .
\end{equation}
They also diagonalize ${\bf B}$, which we see by left-multiplying (\ref{generalized-eigenvalue-equation}) with ${\bf V}^\top$ and right-multiplying with ${\boldsymbol \Lambda}^{-1}$. This yields:
\begin{equation}
{\bf V}^\top {\bf B} {\bf V}  = {\bf V}^\top {\bf A} {\bf V}  {\boldsymbol \Lambda}^{-1} = {\bf I} .
\end{equation}
Thus, the eigenvectors that solve (\ref{generalized-eigenvalue-equation}) jointly diagonalize ${\bf A}$ and ${\bf B}$. The reverse is also true. A matrix ${\bf V}$ that satisfies:
\begin{eqnarray}
\label{diagonalize-A}
{\bf V}^\top {\bf A} {\bf V}  &= {\boldsymbol \Lambda}_A \\
\label{diagonalize-B}
{\bf V}^\top {\bf B} {\bf V}  &= {\boldsymbol \Lambda}_B ,
\end{eqnarray}
where ${\boldsymbol \Lambda}_A$ and ${\boldsymbol \Lambda}_B$ are diagonal matrices, also satisfy (\ref{generalized-eigenvalue-equation}). To see this, solve (\ref{diagonalize-B}) for ${\bf V}^\top$ and substitute this into (\ref{diagonalize-A}), which yields:
\begin{equation}
{\boldsymbol \Lambda}_B {\bf V}^{-1}{\bf B}^{-1}  {\bf A} {\bf V}  = {\boldsymbol \Lambda}_A ,
\end{equation}
and therefore 
\begin{equation}
{\bf B}^{-1}  {\bf A} {\bf V}  = {\bf V} {\boldsymbol \Lambda}_B^{-1} {\boldsymbol \Lambda}_A ,
\end{equation}
which is the same as the eigenvalue equation (\ref{generalized-eigenvalue-equation}) with 
\begin{equation}
{\boldsymbol \Lambda} = {\boldsymbol \Lambda}_B^{-1}{\boldsymbol \Lambda}_A .
\end{equation}

\subsection{Linear combinations}
\label{apdx:linear-combinations}

A matrix ${\bf V}$ that simultaneously diagonalizes two matrices ${\bf A}$ and ${\bf B}$, as in (\ref{diagonalize-A})-(\ref{diagonalize-B}), will also diagonalize any linear combination of these two matrices \citep{fukunaga2013introduction}, because
\begin{eqnarray}
{\bf V}^\top (\alpha {\bf A} + \beta {\bf B}) {\bf V} 
&=&
\alpha {\bf V}^\top{\bf A} {\bf V}  + \beta {\bf V}^\top{\bf B} {\bf V} 
\nonumber
\\
&=&  
\alpha {\boldsymbol \Lambda}_A + \beta {\boldsymbol \Lambda}_B , 
\label{diagonalize-linear-combination}
\end{eqnarray}
which is obviously a diagonal matrix. Therefore, the eigenvectors of ${\bf A}^{-1}{\bf B}$ are also eigenvectors of $(\alpha {\bf A}+ \beta {\bf B})^{-1}(\gamma {\bf A}+ \delta {\bf B})$, assuming vectors $[\alpha, \beta]$ and $[\gamma, \delta]$ are not collinear.  

\subsection{Identifiability}
\label{apdx:identifiability}

An additional important observation is that the eigenvectors of (\ref{generalized-eigenvalue-equation}) are only uniquely defined for unique eigenvalues. If an eigenvalue is degenerate, say with multiplicity $K$, then the corresponding eigenvectors span a $K$-dimensional subspace. Any vector within that subspace is a solution to the eigenvalue equation with that eigenvalue. In this space any arbitrary rotation will yield a valid solutions. Thus, when estimating components, it is possible that solutions with close-by ISC are ``mixed''. Such components are not uniquely identifiable. One should also note that the sign of any eigenvector is arbitrary. If ${\bf v}_d$ is a solution, then so is $-{\bf v}_d$. One can always arbitrarily swap the sign of ${\bf v}_d$ for any component $d$, which also reverses the sign of $y_d^l$ and ${\bf a}_d$ --- the $d$th column of the forward model ${\bf A}$. Thus, keep in mind that in all figures the sign of individual ${\bf v}_d$ and ${\bf a}_d$ are arbitrary.  The same problem arises with most signal decomposition methods such as PCA, ICA or CCA. A common approach used in practice to ``fix'' the sign is to select one sensor/dimension with the strongest forward-model coefficient, and set that to a positive value (if indeed it is significantly larger than other coefficients with opposite sign).

\subsection{Optimization of sum ISC}
\label{apdx:max-sum-isc}

The optimization criterion (\ref{summed-isc}) has the form
\begin{equation}
J({\bf V}) = 
\sum_{d} \frac
{{\bf v}_d^\top {\bf A} {\bf v}_d}
{{\bf v}_d^\top {\bf B} {\bf v}_d}
\end{equation}
and is subject to the constraints that components are uncorrelated (relative to ${\bf B}$):
\begin{equation} 
{\bf v}_d^\top {\bf B} {\bf v}_c =0, \text{for } d \ne c.
\end{equation}
It will be convenient to convert this optimality criterion into conventional Rayleigh quotients. To do so use the Cholesky decomposition (\ref{cholesky-factorization}) as before and replace, ${\bf v}_d = {\bf L}^{-\top} {\bf w}_d$ and ${\bf A} = {\bf L}{\bf M}{\bf L}^\top$ to yield: 
\begin{equation}
J({\bf W}) = 
\sum_{d} \frac
{{\bf w}_d^\top {\bf M} {\bf w}_d}
{{\bf w}_d^\top {\bf w}_d}
\end{equation}
with the constraint now being a conventional orthogonality constraint:
\begin{equation}
{\bf w}_d^\top {\bf w}_c =0, \text{for } d \ne c.
\label{orthogonality-constraint}
\end{equation}
One can now optimize for matrix ${\bf W}$ with the ${\bf w}_d$ as its column vectors. The solution to this constrained optimization problem is well established \citep[][Corollary 4.3.39]{horn1990matrix}. It is given by, ${\bf W}={\bf U}$, the eigenvalue matrix of equation (\ref{eigenvalue-equation}). Inserting ${\bf U}={\bf L} {\bf V}$ into (\ref{eigenvalue-equation}) gives the solution to the original problem, which is given by the generalize value equation (\ref{generalized-eigenvalue-equation}).

To see that the eigenvector matrix of ${\bf M}$ maximizes $J({\bf W})$, first note that except for the constraint, the terms in the sum can be optimized independently, and that they all have the same form. One can therefore consider the maximum of a single term, written here simply as
\begin{equation}
\rho = \frac{{\bf w}^\top {\bf M} {\bf w}}{{\bf w}^\top {\bf w}}  ,
\label{Rayleigh-quotient}
\end{equation}
which is to be optimized subject to the orthogonality constraint (\ref{orthogonality-constraint}). From here on the argument proceeds conventionally. Let ${\bf u}_i, i=1 \ldots D$ be the orthonormal eigenvectors of matrix ${\bf M}$ with eigenvalues $\lambda_1 \ge \lambda_2 \ge \ldots \ge \lambda_D$. We can express the solution vectors in this orthogonal basis as
\begin{equation}
{\bf w} = \sum_{i=1}^D \alpha_{i} {\bf u}_i ,
\label{eigenvalue-basis}
\end{equation} 
where $\alpha_i$ is the projection of ${\bf w}$ on that orthonomal basis of these eigenvectors, $\alpha_i = {\bf u}_i^T {\bf w}$.  Inserting (\ref{eigenvalue-basis}) into (\ref{Rayleigh-quotient}) gives 
\begin{equation}
\rho = 
\frac
{ \sum_i \sum_j \alpha_i {\bf u}_i^\top {\bf M} {\bf u}_j \alpha_j }
{ \sum_i \sum_j \alpha_i {\bf u}_i^\top         {\bf u}_j \alpha_j }
=
\frac{ \sum_i \lambda_i \alpha_i^2}{\sum_i  \alpha_i^2 }  \le 
\frac{ \lambda_{1}  \sum_i \alpha_i^2}{\sum_i  \alpha_i^2 }  =
\lambda_{1}  
.
\label{weighted-eigenvalue-average}
\end{equation} 
This upper bound is attained for $\alpha_1>0$ and all others equal zero, and thus ${\bf w}_1={\bf u}_1$. For the second vector we require that it is orthogonal to the first, and thus we have to constrain $\alpha_1=0$. The sum in (\ref{weighted-eigenvalue-average}) now excludes the first dimension and has therefore an upper bound of $\lambda_2$ which is attained for ${\bf w}_2={\bf u}_2$. The argument continues iteratively until for the last component only, ${\bf w}_D={\bf u}_D$, is orthogonal to all other projections. It also happens to be the minimum of $\rho$.  
\\

\subsection{Uncorrelated components vs orthogonal projections}
\label{apdx:uncorrelated-vs-orthogonal}

We have required here that the extracted components are uncorrelated, i.e. the projection vectors diagonalize ${\bf R}_W$: ${\bf V}^\top {\bf R}_W {\bf V} = {\bf I}$. This means that the component dimensions of ${\bf y}_i^l$ are uncorrelated, but only in the averaged over all subject/repeats $l$. Note that for an individual subject, ${\bf y}_i^l$ may not be uncorrelated. The same is true for MCCA as derived in section~\ref{sec:mcca}. Instead of uncorrelated, we could have required that the projection vectors are orthogonal, ${\bf V}^\top {\bf V} = {\bf I}$, i.e. the linear transformation is limited to rotations. To find the orthogonal (rotations) that maximize ISC, one will have to resort to constrained optimization algorithms \cite[e.g.][]{cunningham2015linear}.  A related issues arises in the context of Canonical Correlation Analysis which attempts to find a linear mapping between two datasets such that the components are uncorrelated \citep{yohai1980canonical,de2012least}. Alternatively, the Procrustes problem aims to find a mapping between two data-sets with orthogonal projections \citep{schonemann1966generalized}.  \citep[The later is what is used for ``hyper-alignment'' between subjects in fMRI by][]{haxby2011common}. CCA and Procrustes are related, and in fact one can smoothly titrate between orthogonal and uncorrelated solutions \citep{xu2012regularized}. In our case we favor uncorrelated components, because we want the components to capture independent information. At a minimum, this means the components have to be uncorrelated. In that sense, our method can be seen as a source separation method, similar to Independent Component Analysis \citep{makeig1996independent}, Denoising Source Separation \citep{sarela2005denoising}, or others \citep{parra2003blind}.
One advantage of restricting the transformations to rotations is that euclidean distances in the new space are preserved. This means that differences found in the original sensor space are preserved in the new component space. Another advantage is that all original dimensions contribute equally to the new dimensions. In contrast, the non-orthogonal shearing operation of CorrCA or CCA may discount some dimensions and emphasize others. Both these advantages of orthonormal mappings, however, require that \emph{all} new dimensions are considered. This is typically the case in the applications of hyper-alignment \citep{guntupalli2016model}. For the purposes of identifying dimensions that are reliably reproduced, however, we find in practice that many dimensions are not significantly correlated between repeats, and one will typically want to remove those. Similarly, one will want to dismiss sensors in the original data that are not reproducible across repeats while emphasizing others. Ultimately, however, the application will dictate whether one should enforce uncorrelated components or orthogonal projections. In either case, the goal can remain to maximize ISC, and the solution can be found, either through an eigenvalue question (as we presented here for uncorrelated components) of with constrained optimization \cite[for orthogonal projections, along the lines of ][]{cunningham2015linear}.

\section{Relationship between Scatter matrices and between- and within-subject covariance matrices}
\label{apdx:scatter-vs-covariance}

Comparing Eq. (\ref{between-scatter-matrix}) to (\ref{total-covariance-matrix}) yields:
\begin{equation}
{\bf R}_T = N {\bf S}_B .
\end{equation} 

One can also relate ${\bf R}_W$ to ${\bf S}_T$ as follows.  Expanding the definition (\ref{between-subject-covariance-matrix}) yields (with abbreviated notation for the sums):
\begin{eqnarray}
{\bf R}_W &=& 
  \sum_{il} {\bf x}_i^l {\bf x}_i^{l \top} 
- \sum_{il} {\bf x}_i^l {\bar {\bf x}}_*^{l \top}  
- \sum_{il} {\bar {\bf x}}_*^l {\bf x}_i^{l \top}
+ \sum_{il} {\bar {\bf x}}_*^l {\bar {\bf x}}_*^{l \top}
\nonumber
\\
&=&
    \sum_{il} {\bf x}_i^l {\bf x}_i^{l \top} 
- T \sum_{ l} {\bar {\bf x}}_*^l {\bar {\bf x}}_*^{l \top}  
\nonumber
\\
&=&
    \sum_{il} {\bf x}_i^l {\bf x}_i^{l \top} 
- T \sum_{ l} {\bar {\bf x}}_*^l {\bar {\bf x}}_*^{l \top}  
+ T \sum_{ l} {\bar {\bf x}}_*^* {\bar {\bf x}}_*^{* \top}  
- T \sum_{ l} {\bar {\bf x}}_*^* {\bar {\bf x}}_*^{* \top}  
\nonumber
\\
&=& 
\left( 
    \sum_{il} {\bf x}_i^l {\bf x}_i^{l \top} 
+ T \sum_{ l} {\bar {\bf x}}_*^* {\bar {\bf x}}_*^{* \top} 
\right) 
-
\left( 
    \sum_{il} {\bar {\bf x}}_*^l {\bar {\bf x}}_*^{l \top}  
- T \sum_{ l} {\bar {\bf x}}_*^* {\bar {\bf x}}_*^{* \top}  
\right)
\nonumber
\\
&=&
    \sum_{il} ({\bf x}_i^l - {\bar {\bf x}}_*^*)
              ({\bf x}_i^l - {\bar {\bf x}}_*^*)^\top 
-
    \sum_{il} ({\bar {\bf x}}_*^l - {\bar {\bf x}}_*^*)
              ({\bar {\bf x}}_*^l - {\bar {\bf x}}_*^*)^\top
\nonumber
\\
&=&
{\bf S}_T  - T {\bf S}_M .       
\end{eqnarray} 
Matrix ${\bf S}_M$, defined in (\ref{variance-of-mean}), captures the variance of the mean across subjects and is zero when all subjects have equal mean: ${\bar {\bf x}}_*^l = {\bar {\bf x}}_*^*$ .\footnote{ 
For time domain signals, this is often the case as it is customary to subtract the mean value (constant offset) to make the signals zero-mean. For rating data, an unequal mean indicates rater bias, i.e. a rater gives systematically lower/higher ratings than other raters, and it may be advantageous to remove this bias by equalizing all mean values. Note that, when ${\bf S}_M=0$, the present definition of ISC is equivalent to the classic definition of intra-class correlation (ICC, see Appendix~B). But in general, this matrix is not zero, which is important when generalizing the approach to non-linear transformations. There, enforcing zero mean in ${\bf x}$ does not guarantee zero mean in ${\bf y}$.} 
With this result, scatter and covariance matrices can be related as follows:
\begin{eqnarray}
{\bf S}_B &=& \frac{1}{N} {\bf R}_T = \frac{1}{N} ({\bf R}_W + {\bf R}_B) , \\
{\bf S}_W &=& {\bf S}_T - {\bf S}_B  
= {\bf R}_W + T {\bf S}_M - {\bf S}_B 
\nonumber
\\
& &
= \frac{N-1}{N} {\bf R}_W - \frac{1}{N}{\bf R }_B   + T {\bf S}_M .   
\end{eqnarray}

\section{Regularization and robustness}
\label{sec:regularization}

Note that a high inter-subject correlation corresponds to large eigenvalues of $\mathbf{R}_{W}^{-1} \mathbf{R}_{B}$. These eigenvalues will be large if the selected direction exhibits large power in the space of $\mathbf{R}_{B}$, or \emph{low} power in the space of $ \mathbf{R}_{W}$.  The former is meaningful and desired, but the latter can occur with rank-deficient or impoverished data and may lead to spuriously high inter-subject correlation.   Truncating the eigenvalue spectrum of $\mathbf{R}_W$ has the desired effect of forcing the extracted data dimensions to have both high power in $\mathbf{R}_B$ \emph{and} $\mathbf{R}_W$.  This makes it less likely to find spuriously reliable dimensions.   Formally, the TSVD approach is to select $\mathbf{V}$ according to:
\begin{eqnarray}
\label{eqn:tsvd}
{\bf R}_B {\bf V} =\tilde{{\bf R}}_W {\bf V}  {\boldsymbol \Lambda},
\end{eqnarray}
where the within-subject covariance is now regularized with the following approximation:
\begin{eqnarray}
\tilde{{\bf R}}_W = \tilde{{\bf U}}_{W} \tilde{{\boldsymbol \Lambda}}_{W}  \tilde{{\bf U}}_{W}^\top,
\end{eqnarray}
where
$\tilde{\bf U}_{W}=\left[ {{\bf u}}_1  \ldots  {{\bf u}}_K  \right]$,  and 
$\tilde{\boldsymbol \Lambda}_{W} = \mathtt{diag} \left[ \lambda_1 \ldots \lambda_K \right] $
are the $K$ principal eigenvectors of ${\bf R}_{W}$
and associated eigenvalues.    The inverse of ${\bf R}_{W}$ is computed according to:
\begin{eqnarray}
\tilde{{\bf R}}_W ^{-1}= \tilde{{\bf U}}_{W} \tilde{{\boldsymbol \Lambda}}_{W}^{-1}  \tilde{{\bf U}}_{W}^\top,
\end{eqnarray}
which then left-multiplies (\ref{eqn:tsvd}) and yields the regularized solution to the CorrCA problem.  

The parameter $K$ is the number of eigenvectors to retain in the representation of ${\bf R}_{W}$.  Decreasing this number strengthens the level of regularization, and $K=D$ is equivalent to not regularizing.

Shrinkage regularization operates by a similar principle: dimensions of the data exhibiting low variance (corresponding to low eigenvalues of ${\bf R}_W$) are enriched by incrementing these eigenvalues:  
\begin{eqnarray}
\label{eqn:shr}
\tilde{{\bf R}}_W = {{\bf U}}_{W}  \tilde{\boldsymbol \Lambda}_{W}  {\bf U}_{W}^\top,
\end{eqnarray}
where
$
\tilde{\boldsymbol \Lambda}_{W}= (1-\gamma) {\boldsymbol \Lambda}_{W} + \gamma \bar{\lambda} {\bf I} , 
$
$\gamma$ is the shrinkage parameter with $0 \leq \gamma \leq 1$ , and $\bar{\lambda} = \mathrm{Tr}({\bf R}_W) / D $ is the mean eigenvalue of the (unregularized) within-subjects covariance. The effect of shrinkage, which retains the full dimensionality of the covariance matrix, is that the smaller eigenvalues are increased and larger eigenvalues are reduced (hence the term ``shrinkage'').  

The advantage of shrinkage regularization is the simplicity of its implementation, where the regularized covariance can be simply computed as
\begin{equation}
\tilde{{\bf R}}_W = (1-\gamma) {\bf R}_W + \gamma \bar{\lambda} {\bf I} 
\end{equation}
without any further modifications to the generalized eigenvalue routine. TSVD instead requires more careful handling of the null-space when computing the general eigenvalues (see provided code). On the other hand, TSVD has the important advantage that the solutions ${\bf V}$ to the regularized eigenvalue problem (\ref{eqn:tsvd}) diagonalize $\tilde{\bf R}_W$ as well as the original ${\bf R}_W$. This means that the extracted components are strictly uncorrelated, whereas for shrinkage that is only approximately the case.

The performance of TSVD and shrinkage can be evaluated by maximizing ISC on one data set (training set) and using the resulting projection vectors to measure ISC on a different data set (test set). We do this on one of the application examples in section~\ref{app:regularization-evaluation}. 

Note that the TSVD approach can also be used to regularize the MCCA algorithm. In this case, the regularized inverse $\tilde{\bf D}$ has to be computed separately for each block of the block-diagonal matrix ${\bf D}$ in (\ref{mcca-block-matrizes}).

\section{Testing for statistical significance}
\label{sec:significance}

The assessment of statistical significance of $\rho$ requires knowledge of their distribution under the null hypothesis. The null hypothesis associated with CorrCA states that there exists no shared one-dimensional subspace in which the data of at least two subjects are correlated. Estimating the null distribution is complicated by two factors. First, CorrCA has free parameters, the optimization of which inflates the ISC values under the null hypothesis to an extent governed by the dimensionality of the problem (parameters $T$, $D$, and $N$). Second, temporal correlations in the data lead to larger variances than expected under the assumption of independent and identically distributed (IID) data typical for parametric statistical tests, and, thereby, to false positive detections of significant ISC values. The IID assumption may be reasonable for rating data, where rated individuals may be independent from one another, but it is not fulfilled for time series data such as EEG recordings. 
Here, we present three approaches to estimate the significance of individual correlated components as well as the dimensionality of the correlated subspace for IID as well as auto-correlated data.

{\sl Parametric null distribution}---For IID data, we can use the F-statistics given in Eq. (\ref{F-statistic}) to test for significant ISC (provided the means are equalized). Evidently, this statistic is not applicable for $\rho$ values obtained on training data, as the we have adapted parameters to optimize this statistic itself. However, we can use the optimal projection vectors to assess statistical significance on separate, previously unseen test data. An important limitation is that the $T$ exemplars have to be independent. This is true perhaps for rating data, but typically not true for temporal signals, where samples in time are often correlated. For such data, the non-parametric test discussed above are more appropriate. To avoid overestimating the number of components due to multiple testing (there are always $D$ components to be tested), we adopt a Bonferroni correction. This  limits the family-wise error rate, i.e. the probability of making one or more false discoveries among the $D$ tests. Thus, only $S$ values with $p<\alpha/D$ are considered significant, where $\alpha$ is the desired significance level and $p$ is computed using the F-statistics.

{\sl Phase-scrambled surrogate data}--- For auto-correlated data we have to use shuffle statistics. This approach was introduced by \cite{theiler1992testing} as a tool to test for nonlinearities in time series data.
The idea is to create data that are consistent with the null hypothesis but otherwise resemble the observed experimental data in terms of temporal and spatial correlations. The original surrogate data of \cite{theiler1992testing} preserves the amplitude spectrum of the original data but uses a random phase spectrum, which is achieved through an application of the Fourier transform and its inverse. This approach was extended to additionally maintain the spatial covariance structure of multivariate data by using identical random phase shifts for all variables by \cite{prichard1994generating}. Here, we use the approach of \citeauthor{prichard1994generating} to test for significant $\rho$ based on the consideration that no inter-subject correlation can be present after phase-scrambling the data of each subject. Variants of this approach have previously been used to test for significant correlations \citep{schaworonkow2015power, ki2016attention, cohen2016memorable, Haufe207456}. The p-value for the ISC of a given correlated component is defined as the frequency with which the ISC computed from the random-phase surrogate data is exceeded by the maximum ISC of the most strongly correlated component in the
original data. Note that for every shuffle we obtain a set of ISC values, $\rho_d$, for the $d$th component. When we compare the ISC value on the original data, we measure p-value based on the distribution of maximum $\rho$, i.e. all $\rho_d$ values are compared against the Null distribution of $\rho_1$. With this approach no further correction for multiple comparison is needed.

{\sl Random circular shifts}---A simpler approach to create surrogate data for auto-correlated data is to circularly shift the samples of each subject by a different offset, where the same offset is used for all variables of the same subject (i.e. sample index $i=1 \ldots T$ are shifted by a random offset $o$: $i \leftarrow (i+ o) \text{mod } T + 1 $). Just as the phase-scrambling approach, this procedure ensures that spatial correlations and spectral properties of the original data are maintained. Additionally, non-linear properties of the original data are maintained. This approach has previously been used in \cite{dmochowski2012correlated,dmochowski2014audience}. P-values are defined analogous to the phase-scrambling approach.

We test the validity of the three methods on simulated data in section~\ref{sec:simulations}. The goal is to determine under which conditions these methods determine the correct number of underlying correlated components.

%Here, we use that Fisher-transformed Pearson correlations $\text{arctanh}(r)$ obtained from two IID Gaussian samples of length $T$ are approximately normal distributed with variance $1/(T-3)$\citep{fisher1915frequency}. Analogously, a normal distribution with variance $\text{Var}[z] = 1/(NT/2-3)$ can be assumed for $z = \text{arctanh}(\rho)$, where the additional factor $N/2$ accounts for the presence of more than two subjects. P-values can be derived by evaluating the cumulative distribution function $\Phi$ of the standard normal distribution: $p = \Phi(-z / \surd \text{Var}[z])$. \LP{Does this account for the training effect? If so, why does it not depend on number of free parameters $D$?}

\section{Validation on simulated data: Data generation and evaluation methods}
\label{sec:simulation-evaluation-methods}

The generated signal and noise components, ${\bf s}^l(t) \in \mathbb{R}^K$ and ${\bf n}^l(t) \in \mathbb{R}^D$, $t = 1, \hdots, T$, $l=1, \hdots, N$, were mapped to the measurement space as ${\bf x}_\text{s}^l(t) = {\bf A}_\text{s}^l {\bf s}^l(t)$, ${\bf x}_\text{n}^l(t) = {\bf A}_\text{n}^l {\bf n}^l(t)$ and normalized as ${\bf x}_\text{s}^l(t) \leftarrow {\bf x}_\text{s}^l(t) / ||{\bf x}_\text{s}^l||_F$, ${\bf x}_\text{n}^l(t) \leftarrow {\bf x}_\text{n}^l(t) / ||{\bf x}_\text{n}^l||_F$, where $||{\bf x}||_F$ denotes the Frobenius norm of the multivariate data set ${\bf x}(t)$. The mixing matrices ${\bf A}_\text{s}^l = {\bf O}_\text{s}^l {\bf D}_\text{s}^l \in \mathbb{R}^{D \times K}$ and ${\bf A}_\text{n}^l = {\bf O}_\text{n}^l {\bf D}_\text{n}^l \in \mathbb{R}^{D \times D}$ were generated as products of matrices ${\bf O}_\text{s}^l \in \mathbb{R}^{D \times K}$ (${\bf O}_\text{n}^l \in \mathbb{R}^{D \times D}$) with random orthonormal columns and diagonal matrices ${\bf D}_\text{s}^l \in \mathbb{R}^{K \times K}$ (${\bf D}_\text{n}^l \in \mathbb{R}^{D \times D}$), the diagonal entries of which were drawn randomly as ${{D}}_{ii} = \exp (d_i)$, $d_i \sim {\mathcal N}(0, 1)$ and normalized to a maximum of 1. This ensured that the non-zero eigenvalues of the signal and noise covariances matrices decayed exponentially as is the case for many real-world data sets. Unless otherwise noted, the same mixing matrices ${\bf A}_\text{s}^l \equiv {\bf A}_\text{s}$ and ${\bf A}_\text{n}^l \equiv {\bf A}_\text{n}$ were used for all subjects, reflecting the fundamental assumption of CorrCA that the spatial covariance structure of signal and noise components is the same in all subjects. The emulated data were generated as ${\bf x}^l(t) = \xi {\bf x}_\text{s}^l(t) + (1-\xi) {\bf x}_\text{n}^l(t)$, where the scalar parameter $\xi~=~10^{\nicefrac{\text{SNR}}{20}} / (1 + 10^{\nicefrac{\text{SNR}}{20}})$ was used to adjust the SNR. Note that SNR was defined here in terms of the contribution of the signal and noise components to the Frobenius norm of the entire multivariate measurement. It is, therefore, different from the definition provided in Eqs.~\eqref{Separation} and \eqref{SNR}, which applies to single channels or components and relates to the ISC through Eq.~\eqref{ISCtoSNR}. Note that in most of our experiments, the ISC of each simulated signal component was set to 1, corresponding to infinite SNR according to Eq.~\eqref{ISCtoSNR}.

We applied CorrCA to the emulated data and measured its performance in terms of the achieved ISC (average over the first $K$ components) on the training set. We also measured the average ISC for a test set, i.e., using the same CorrCA projections ${\bf V}$ on new emulated data of identical size. Moreover, we measured the performance of CorrCA to reconstruct the true (emulated) correlated components ${\bf s}^l(t)$ and their corresponding mixing matrices ${\bf A}_\text{s}^l$ by comparing them to ${\bf y}^l(t)$ and the estimated forward model ${\bf A}$ (section~\ref{sec:forward-model}). The performance metric for this was the Pearson correlation between individual simulated and reconstructed components or columns of ${\bf A}_\text{s}$. With this, we measure if each component was correctly identified. However, because components with equal ISC can be arbitrarily mixed within a subspace (Appendix~\ref{apdx:identifiability}), we also assessed identifiability in terms of the angle between the subspaces spanned by the first $K$ columns of the estimated $\hat{\bf A}$ and ${\bf A}_\text{n}$, as well as the angle between the simulated components ${\bf s}^l(t)$ and the first $K$ dimensions of ${\bf y}^l(t)$ reconstructed by CorrCA. Angles were normalized to the interval $[0, 1]$. Finally, we estimated the number of simulated correlated components, $K$, using phase-scrambled surrogate data, surrogate data obtained using random circular shifts, and the parametric F-test. Empirical null distributions were obtained using 1000 surrogate data sets. The number of components with p-values smaller than $\alpha = 0.05$ was used as an estimate for $K$. For the parametric test, the $T$ samples were randomly split into training and test sets of equal size, where the CorrCA projections ${\bf V}$ were obtained on the training set, and the statistical assessment was conducted on the test set. For this approach, the median number of components with p-values smaller than $\alpha = 0.05$ across 100 random train/test splits was used as an estimate of $K$.

All experiments were repeated 100 times. Mean values and standard deviations of all performance measures are depicted in Figures~\ref{fig:sim-SNR-dependence}---\ref{fig:sim4}.

\section{Extension to non-linear maps using Kernels}
\label{sec:kernel-corrca}

So far we have only considered linear transformations between ${\bf x}$ and ${\bf y}$. Now, given the close relationship between CorrCA and LDA, we can extend CorrCA to non-linear transformations following the approach of kernel-LDA \citep{mika1999fisher,baudat2000generalized,zhang2004kernel,de2005eigenproblems}. Assume a non-linear mapping ${\boldsymbol \Phi}(\cdot)$ for the data of subject $l$:
\begin{equation}
{\bf Z}^l = {\boldsymbol \Phi}({\bf X}^l) .
\end{equation}
In this notation, ${\bf X}^l$ has dimension $D \times T$ spanning the entire multivariate time series for subject $l$, and ${\bf Z}^l$ has dimensions $D' \times T'$. Thus, the non-linear transformation ${\boldsymbol \Phi}(\cdot)$ is a mapping from space-time to a new space with possibly quite different dimensions. 
\footnote{In the example of brain signals, ${\bf Z}^l$ could represent, for instance, the power spectrum in source space \citep{michel2004eeg}, where $D'$ captures the number of sources and $T'$ the number of frequency bins. In the application examples and in the code we simplify this by applying the non-linear mapping to each time point separately, so that $T=T'$ and we have a purely spatial nonlinear mapping, similar to conventional kernel-LDA. However, this more general formulation allows one in principle to reduce the time dimension as well. This is important because the computational complexity of the kernel approach scales with $O(T^3)$, and thus becomes prohibitive for long time-domain signals.} 
The approach now is to first extract non-linear features of the data with ${\boldsymbol \Phi}(\cdot)$, and then combine these features linearly with projection ${\bf v}$: 
\begin{equation}
{\bf y}^l = {\bf v}^\top {\bf Z}^l.
\label{nonlinear-projection}
\end{equation}
Note that the transformation ${\boldsymbol \Phi}(\cdot)$ does not need to be defined explicitly.  Instead, the ``kernel trick'' \citep{scholkopf1998nonlinear} allows one to specify this non-linearity only implicitly, by defining instead the inner product of vectors in this new $D'$-dimensional feature space with a kernel function ${\bf K}( \cdot , \cdot )$:
\begin{equation}
{{\bf Z}^k}^\top {\bf Z}^l = {\bf K}({\bf X}^k, {\bf X}^l) . 
\end{equation}  
Knowing how to compute the inner product is sufficient, if all the expressions in the algorithm can be formulated in terms of this inner product.\footnote{Note that this inner product is only over dimension $D'$. The function ${\bf K}( \cdot , \cdot )$ is therefore only a "kernel" in this $D'$ dimensional space.} In a slight abuse of notation, we will write ${\bf K}^{kl} = {\bf K}({\bf X}^k, {\bf X}^l)$ as shorthand for this matrix defined for each pair of subjects $lk$. Note that the dimensions of ${\bf K}^{kl}$ are $T' \times T'$ and thus this matrix can potentially be quite large. In cases where the non-linear transformation ${\boldsymbol \Phi}(\cdot)$ can be expressed explicitly, and $D' \ll T'$, one may chose to operate directly with ${\bf Z}^l$. Otherwise, it will be beneficial to leverage the definition of the non-linear transformation in terms of ${\bf K}^{kl}$. In the following, we will rewrite the optimization criterion $\rho$ in terms of ${\bf K}^{kl}$ alone.  

\begin{figure}[htbp]
\includegraphics[width=\columnwidth]{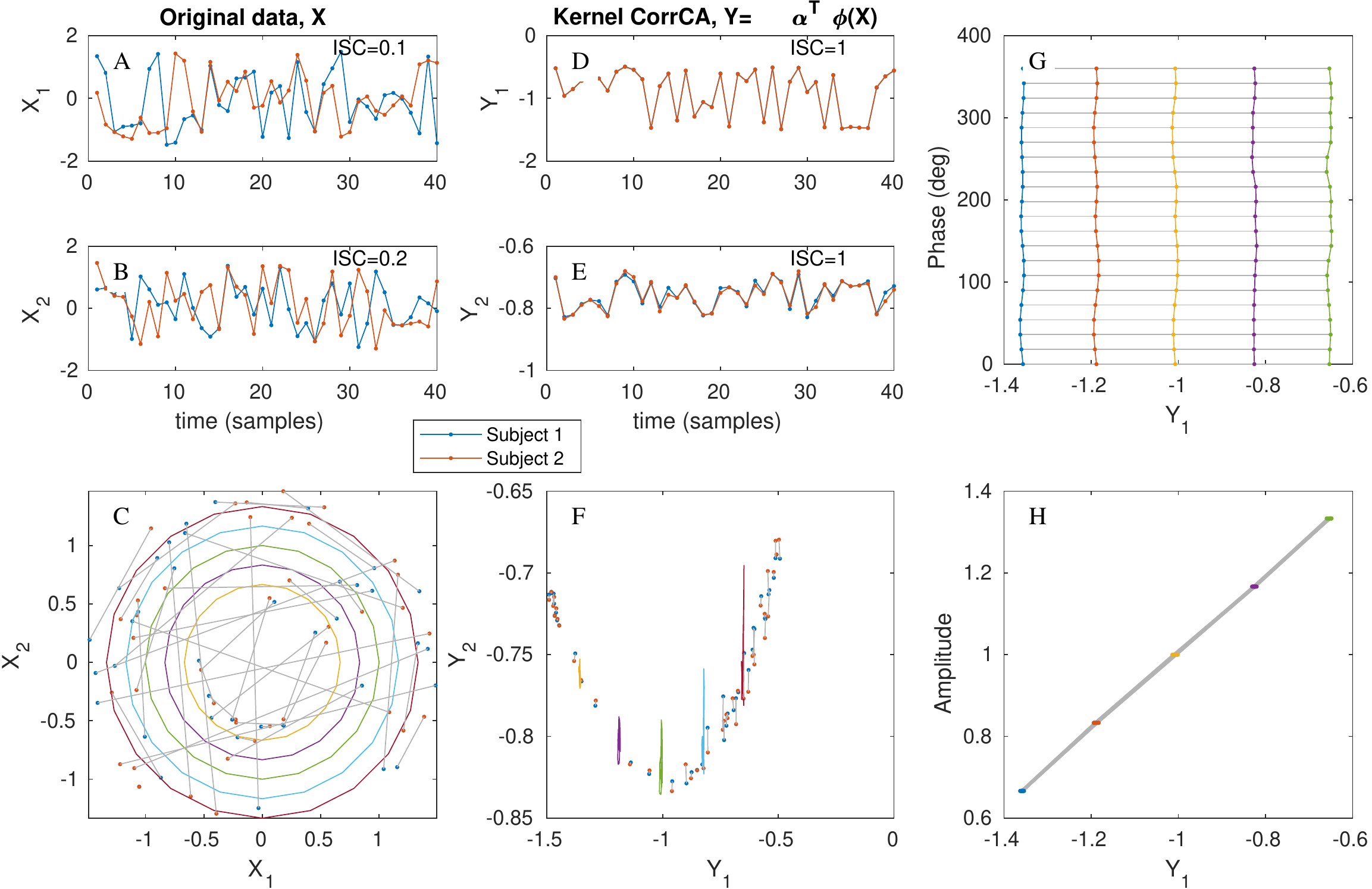}
\caption{{\bf Kernel-CorrCA.} In this demonstrative 2D example, two subjects share amplitude but have a random phase. The goal is for the algorithm to discover the common amplitude signal --- a non-linear transformation of the data --- from 60 samples. (A-B) Two dimensions of the original data with one trace for each subject. (C) Samples are connected between the two subjects with gray lines, as in Figure~\ref{fig:toy-example-2D}. Circles of varying radii are drawn for visualization purposes. Note that connected points have the same distance from the centers, i.e. subjects share an amplitude. (D-E) After kernel-CorrCA using Gaussian kernels, signals between subjects are strongly correlated (ISC=1). (F) Points are in a new non-linear space, which is necessarily curved to decorrelate the two dimensions. (G-H) Colored lines correspond to the circles in panel C. The first dimension $y_1$ is independent of phase and grows linearly with amplitude. This means that the algorithm identified the non-linear dimension that is shared between the two subjects. }
\label{fig:kernel-CCA} 
\end{figure}

The crucial step is to define the projection vectors in terms of the samples of the mean across subjects. 
\begin{equation}
{\bf v} = \sum_{i=1}^{T'} \alpha_i {\bar {\bf z}}^*_i = {\bar {\bf Z}}^* \balpha,
\label{v-mean-model}
\end{equation}
where ${\bar{\bf z} }^*_i$ is the across-subject mean vector for exemplar $i$ in the non-linear mapped space ($i$th column of $\bar{\bf Z}^*$).  Here, $\alpha_i$ are parameters that indicate how exemplars $\bar{\bf z}_i$ are to be combined to represent the projection vector ${\bf v}$. The approach will be to find optimal vector $\balpha$ instead of vector ${\bf v}$, and thus we need to rewrite the optimality criterion in term of $\balpha$.\footnote
{As an alternative to (\ref{v-mean-model}), we could have also defined the projection vector in terms of the samples of all subjects:
\begin{equation}
{\bf v} = \sum_{l=1}^N \sum_{i=1}^{T'} \alpha_i^l {\bf z}^l_i  .
\end{equation}

The derivations can all be repeated analogously and the resulting algorithm is the same, but with $NT'$ degrees for freedom in $\balpha$. In the provided code, this is referred to as the ``full model'', whereas equations derived here for the expansion in terms of the mean (\ref{v-mean-model}) are referred to as the ``mean model''. In the example provided in Figure~\ref{fig:kernel-CCA}, the numerical results are very similar for both models. The question as to whether the mean model is sufficient is empirical and comes down to whether there is enough diversity in the mean response to capture the individual variations across subjects. Mathematically, this implies that the expansion in terms of the mean is complete, i.e., the mean response matrix ${\bar {\bf Z}^*}$ is full rank.
} 

Combining (\ref{nonlinear-projection}) with (\ref{v-mean-model}) one can now express the non-linear components ${\bf y}$ as a linear combination of ${\bf K}^{lk}$ as follows:
\begin{eqnarray}
{\bf y}^l &=& {\bf v}^\top {\bf Z}^l 
=   
\balpha^\top \frac{1}{N} \sum_{k=1}^N {\bf Z}^{k^\top} {\bf Z}^l
\nonumber
\\
&=&
\balpha^\top \frac{1}{N} \sum_{k=1}^N {\bf K}^{kl}
=
\balpha^\top {\bar {\bf K}}^{*l} ,
\end{eqnarray}
where the bar in ${\bar {\bf K}}^{*l}$, indicates that we are taking the mean, and the asterisk specifies over which index this mean is taken. It analogously follows that,
\begin{equation}
{\bf v}^\top {\bar {\bf Z}}^* = \balpha^\top {\bar {\bf K}}^{**} .
\end{equation}
Denote the columns of matrix ${\bf Z}^l$ and ${\bar {\bf K}}^{*l}$ as ${\bf z}_i^l$ and ${\bar {\bf k}}_i^{*l}$. Then we can also write with this notation for the average:
\begin{equation}
{\bf v}^\top {\bar {\bf z}}_i^* = \balpha^\top {\bar {\bf k}}_i^{**} .
\end{equation}
The within-subject and total variance of the non-linear features ${\bf z}_i^l$ can thus be written as:
\begin{eqnarray} 
r_W &=& {\bf v}^\top {\bf R}_W {\bf v}
=
\sum_{l=1}^{N} \sum_{i=1}^{T'}   
{\bf v}^\top 
( {\bf z}_i^l - {\bar {\bf z}}_*^l )
( {\bf z}_i^l - {\bar {\bf z}}_*^l )^\top 
{\bf v} 
\nonumber
\\
&=&
\balpha^\top 
{\bf C}_W
\balpha , \\
r_T &=& {\bf v}^\top {\bf R}_T {\bf v}
=
N^2 \sum_{i=1}^{T'} 
{\bf v}^\top 
( {\bar {\bf z}}_i^* - {\bar {\bf z}}_*^* )
( {\bar {\bf z}}_i^* - {\bar {\bf z}}_*^* )^\top 
{\bf v} 
\nonumber
\\
&=&
\balpha^\top 
{\bf C}_T
\balpha ,
\end{eqnarray}
where we defined the within-subject and total covariance of ${\bf k}_i^{*l}$ as:
\begin{eqnarray}
{\bf C}_W &=& 
\sum_{l=1}^{N} \sum_{i=1}^{T'} 
( {\bar {\bf k}}_i^{*l} - {\bar {\bf k}}_*^{*l} )
( {\bar {\bf k}}_i^{*l} - {\bar {\bf k}}_*^{*l} )^\top , \\ 
{\bf C}_T &=& 
N^2 \sum_{i=1}^{T'} 
( {\bar {\bf k}}_i^{**} - {\bar {\bf k}}_*^{**} )
( {\bar {\bf k}}_i^{**} - {\bar {\bf k}}_*^{**} )^\top . 
\end{eqnarray}
Because of the symmetry of these expressions with the definitions of ${\bf R}_W$ and ${\bf R}_T$, we have again, ${\bf C}_B = {\bf C}_T - {\bf C}_W$, and can therefore write the inter-subject correlation in the non-linear space as a function of the new parameters $\balpha$ as follows:
\begin{equation}
\rho = 
\frac{{\bf v}^\top {\bf R}_B {\bf v}}{{\bf v}^\top {\bf R}_W {\bf v}} 
=
\frac{\balpha^\top {\bf C}_B \balpha}{\balpha^\top {\bf C}_W \balpha}.
\end{equation}
This can be solved again as an eigenvalue problem but now with the within- and between-subject covariance matrices of the kernel vectors $\bar{\bf k}_i^{*l}$ and the corresponding projection vector $\balpha$.  \edit{Note that a similar result is obtained for the kernel version of CCA, namely, the conventional CCA algorithm is applied to kernel matrices instead of the original data \citep{lai2000kernel}.}

To demonstrate the method, we generated a simple 2-dimensional example where the shared dimension is non-linear, and this non-linear relationship has to be discovered from the data alone. As with the example for linear CorrCA (Figure~\ref{fig:toy-example-2D}), we have simulated $N=2$ ``subjects'', each providing two signals ($D=2$), now with $T=40$ samples in time (Figure~\ref{fig:kernel-CCA}~A~\&~B). In order to test whether the algorithm finds the correct non-linear mapping we generate data with a known non-linear relationship. In this case the two subjects share the amplitude (distance from the origin in the 2-dimensional input space of ${\bf x}$; see Figure~\ref{fig:kernel-CCA}~C). However, the phase angle in this 2-dimensional plane is selected randomly at each time point and for each subject. We apply kernel-CorrCA using Gaussian kernels \citep{zhang2004kernel} keeping the time axis unchanged ($T=T'$; see also our companying code). The first component dimension found by kernel-CorrCA, $y_1$, is approximately linear with the amplitude and is independent of phase in the original 2D plane (Figure~\ref{fig:kernel-CCA}~G~\&~H). The algorithm therefore discovered in the first component amplitude as the shared non-linear dimension. The second component dimension, $y_2$ is essentially a nonlinear function of component $y_1$ (Figure~\ref{fig:kernel-CCA}~F), hence it also has high ISC. The specific non-linearity between the two components is arbitrary, but does enforce that the two variables are uncorrelated. We obtain nearly identical results regardless of whether we use Gaussian or Tanh kernels \citep[see code and][]{zhang2004kernel}, although the specific relationship between $y_1$ and $y_2$ differs.

\section{Further remarks on Inter-Subject Correlation}

\subsection{Relationship of ISC to Intra-Class Correlation}
\label{apdx:ICC}

The definition of inter-subject correlation in the present work is similar to the definition of intra-class correlation (ICC) first introduced by \cite{fisher1925statistical}. Fisher's ICC is referred to as pairwise inter-class correlation and differs from Pearson's correlation coefficient \citep{donner1986review}. A more modern definition of intra-class correlation, which in the present context would allow for a different number of subjects per sample (i.e. $N_i$), was introduced by \cite{karlin1981sibling}:
\begin{equation}
r_{p} = \frac{\sum_{i=1}^{T} \frac{1}{N_i-1} \sum_{k=1}^{N_i} \sum_{l=1, k \neq l}^{N_i} (y_i^l - {\bar y}_*^*)(y_i^k - {\bar y}_*^*)}
{\sum_{i=1}^{T} \sum_{l=1}^{N_i} (y_i^l - {\bar y}_*^*)^2}.
\end{equation}

The definition of pairwise ICC, $r_p$, differs from our definition of ISC, $\rho$, in that it subtracts the total  mean $\bar{y}_*^*$, whereas we subtract the sample mean of each subject, $\bar{y}_*^l$. The two are identical in the case of unbiased ratings, i.e. ${\bf S}_M=0$ defined in Eq.~\eqref{variance-of-mean}. A definition comparable to ISC can be found in \citep{mckeon1966canonical} in the context of MCCA under the term ``modified inter-class correlation'', but is not commonly seen in the literature. Instead, the generalized eigenvalue solution to MCCA presented here has been often derived using the sum of correlations, subject to a normalization constraint \cite[e.g.][]{nielsen2002multiset,via2005canonical,asendorf2015informative}.

In Table~\ref{table-indice}  we summarize various uses of these correlation measures. For each case the indices play different roles. In the context of ICC, index $i$ represents different classes whereas $l,k$ are different exemplars of a class. Each class can have a different number of exemplars $N_i$. In the context of ISC, $l, k$ are subjects and $i$ represent measures taken, perhaps in time. In the case of ratings, $l,k$ are the raters and $i$ are the items being rated. In \citet{karlin1981sibling}, the measure $r_p$ is used to assess intra-familial correlations (correlations between siblings). In this case, $k,l$ indexes the siblings, whereas $i$ enumerates the families. The algorithms listed in Table~\ref{table-indice} aim to extract components in multi-dimensional data that maximize the corresponding statistic. Where no algorithm is listed, either CorrCA or LDA could be used.

\begin{table}
\caption{Meaning of different indices used in the definitions of various measures of correlation. }
\label{table-indice}
\begin{center}
\begin{tabular}{lllll}
\headrow
\thead{Algorithm} & \thead{Statistic}   & \thead{i=1 \dots T} & \thead{l,k=1 \dots N}        & \thead{Reference}
\\ 
CorrCA    & inter-subject correlation  & samples & subjects  & present paper
\\ 
JD		  & Signal-to noise ratio      & samples & repeats   & \citep{de2014joint}  
\\ 
LDA       & between class separation   & classes & exemplars & \citep{rao1948utilization}
\\ 
          & intra-class correlation    & classes & exemplars & \citep{donner1986review}
\\ 
          & intra-rater correlation    & items   & raters    & \citep{rousson2002assessing}
\\ 
          & intra-familial correlation & families& siblings  & \citep{karlin1981sibling} 
\\ 
MCCA      & modified intra-class correlation & samples & datasets  & \citep{mckeon1966canonical} 
\\ 
MCCA      & inter-subject correlation  & samples & subjects  & present paper 
\\ \hline
\end{tabular}
\end{center}
\end{table}

The definition of ISC/IRC in Eq. (\ref{isc}) can be thought of as a Pearson correlation coefficient, where we concatenate the signals (after subtracting the individual subject/rater means) corresponding to all possible pairs of subjects. In contrast to conventional Pearson correlation, with this symmetric definition (where we include pairs $lk$ as well as $kl$), the signals of different subjects have to have identical scale in order to achieve perfect ISC/IRC (see section~\ref{apdx:normalization-of-isc}). In the case of ratings, this means that different raters can only differ in their mean ratings but have to otherwise provide identical deviations from the mean in order to achieve perfect inter-rater correlation. This means that this metric is sensitive to multiplicative bias (i.e. each rater/subject has a different scale). The issue of bias in the context of inter-rater agreement is discussed in more detail by \cite{rousson2002assessing}, who argue that Pearson correlation should be used to assess test-retest reliability to account for learning effects of subjects from one test to the next. If there is an undesirable proportional bias, then one can easily correct for this bias by standardizing the data, i.e. dividing by the standard deviation for each subject/rater in each dimension prior to applying CorrCA. 

As we have shown here, maximizing IRC is equivalent to maximizing inter-rater-reliability, where we have defined reliability in Eq. (\ref{SNR}) as the variance of the mean over the mean of the variance. For instance, if the rating is the speed of performing a task (as in the application \ref{app:ratings}), this would mean that a 100 millisecond variation across repeats is fairly reliable for a task that takes 5 seconds to complete, but really quite unreliable if the task only takes 0.5 second in average. 

It should be noted that, when ratings are on a discrete numerical scale, IRC may not be an appropriate measure of reliability. For instance, a rating on an integer scale of 1, 2, 3, 4, 5, may have an inter-rater variability of $\pm 1.0$. With the present definition of IRC --- $\rho$ in Eq. (\ref{isc}) --- this same variability would be considered half as reliable if the mean rating is 4 as compared to a mean of 2. Whether this is desirable or not may depend on the specific ratings statistic. Finally, IRC can only be used meaningfully for ratings that are on a proportional scale. For categorical ratings one should use instead the conventional Cohen's $\kappa$ or similar inter-reliability measures.

\subsection{ISC is normalized}
\label{apdx:normalization-of-isc}

To our knowledge, the present definition of $\rho$ in Eq. (\ref{isc}) does not appear in this exact form in the literature, except for our previous work \citep{cohen2016memorable}. Therefore, we take some time here to show that this definition of correlation has indeed a maximal value of 1, and that this occurs only when the all signals are identical across subjects (not just proportional as in Pearson's correlations), except for a subject-dependent mean, which is permitted.

Without loss of generality, let us assume that all signals are zero mean, ${\bar y}^l_*=0$. To show that $\rho \leq 1$, we can also demonstrate, equivalently, that the following expression is non-negative:
\begin{eqnarray}
0 &\leq&  (N-1) r_W - r_B  
\nonumber \\
&=& 
(N-1) \sum_l \sum_i  (y^l_i)^2 - \sum_l \sum_{k\neq l}  \sum_i y^l_i y^k_i 
\nonumber \\
&=& 
N \sum_l \sum_i  (y^l_i)^2 - \sum_i  \left( \sum_l  y^l_i \right)^2 =E .
\label{E-definition}
\end{eqnarray}
The last expression is abbreviated as $E$. To find the $y^l_i$ that minimizes $E$ we find the location where the gradient is zero and check whether the curvature at that location is non-negative. 
\begin{eqnarray}
\frac{\partial E}{\partial y_j^k} 
&=& 
2 N y^k_j - 2 \sum_i   \sum_l  y^l_i \sum_l \delta^{lk}_{ij} 
\nonumber \\
&=& 
2 N y^k_j - 2 \sum_l  y^l_i   ,
\label{E-gradient}
\\
\frac{\partial^2 E}{\partial y_j^k \partial y_j^l} 
&=& 
2N\delta^{kl}_{ji} - 2 \sum_n \delta^{kn}_{ij}
\nonumber \\
&=& 
 2\delta_{ij} (N\delta^{kl}-1).
\label{E-curvature}
\end{eqnarray}
Solving for $\partial E/\partial y=0$  yields an unique solution:
\begin{equation}
y_j^k = \frac{1}{N} \sum_l y_j^l = \bar{y}^*_j.
\label{max-rho-solution}
\end{equation}
This solution is a linear manifold, as one can pick arbitrary scale for $\bar{y}^*_j$. Inserting this into (\ref{E-definition}) yields $E=0$ at that point. To determine whether this is indeed a minimum value for $E$, and hence $\rho \leq 1$, we need to show that the Jacobian (\ref{E-curvature}) has only non-negative eigenvalues. Because of the $\delta_{ij}$ in (\ref{E-curvature}), we can find the eigenvalues for each $ij$ separately, i.e. the eigenvalues of $J = N  {\bf I} - {\bf 1}$, where ${\bf I}$ is the $N$-dimensional identity matrix and ${\bf 1}$ is a $N$-dimensional vector with all values set to 1. This ${\bf J}$ is the same for all $ij$, and the eigenvalue equation is for ${\bf y} \in \mathbb{R}^N$: 
\begin{equation}
{\bf J} {\bf y} = \lambda {\bf y} .
\label{J-eigenvalue}
\end{equation}
The characteristic polynomial for these eigenvalues reads:
\begin{eqnarray}
0&=& \text{det}\left( N{\bf I} - {\bf 1}{\bf 1}^\top - \lambda {\bf I})\right) 
\nonumber \\ 
&=&
(N-\lambda)^N \text{det}( {\bf I} - (N-\lambda)^{-1} {\bf 1}{\bf 1}^\top) 
\nonumber \\ 
&=&
(N-\lambda)^N ( 1 - (N-\lambda)^{-1} N) 
\nonumber \\ 
&=&- (N-\lambda)^{N-1} \lambda .
\end{eqnarray} 
This means that there is a $(N-1)$-fold eigenvalue, $\lambda=N$, and a unique eigenvalue, $\lambda=0$. Therefore, the Jacobian is semidefinite, i.e. it has positive curvature in all directions, except in the direction of the eigenvectors of $\lambda=0$. Inserting this in (\ref{J-eigenvalue}) yields ${\bf y}=c {\bf 1}$ with a free parameter $c$. If we compare that to (\ref{max-rho-solution}), we see that these are the same, with $c=y_i^*$. This means that, when all signals are the same for all subjects, then we achieve a global minimum for $E$ and, therefore, $\rho$ is maximal and equals to 1. The curvature is zero in one direction at the minimum because one can change the global scale of all signals without changing $E=0$.   

\subsection{ISC between individuals and a group of subjects}
\label{apdx:individual-isc}
 
In a number of studies, we compare brain responses of individual subjects to that of the group by correlating between pairs of subjects, and then averaging over pairs involving the subjects of interest. This correlation of individual subject with a group is then used to compare with individual measures \citep{ki2016attention,cohen2016memorable,petroni2018variability}. For this purpose, we used the following definition for single-subject ISC \citep{cohen2016memorable}: 
\begin{eqnarray}
\rho_k &=& \frac
{\sum_{l=1,l \ne k}^{N} (r_{kl}+r_{lk})}
{\sum_{l=1,l \ne k}^{N} (r_{ll}+r_{kk})} ,\\
r_{kl} &=& \sum_{i=1}^{T} (y_i^k - \bar{y}_*^k)(y_i^l - \bar{y}_*^l).
\end{eqnarray}
This can be computed for each extracted component $y_j(t)={\bf v}^\top_j {\bf x}(t)$.

\bibliography{cca-references}

\end{document}